%% file: main.tex
\documentclass{article}

\usepackage[final]{colm2025_conference}
\usepackage{microtype}
\usepackage{hyperref}
\usepackage{booktabs}

\usepackage{lineno}

\definecolor{darkblue}{rgb}{0, 0, 0.5}
\hypersetup{colorlinks=true, citecolor=darkblue, linkcolor=darkblue, urlcolor=darkblue}

\usepackage[T1]{fontenc}    

\input{macros}
\usepackage{multirow}
\usepackage{bbm}
\usepackage{bm}
\usepackage{graphicx}
\usepackage{subcaption}
\usepackage{diagbox}
\usepackage{amsfonts}       
\usepackage{nicefrac}       
\usepackage[dvipsnames]{xcolor}

\title{Mixture of Attention Spans: Optimizing LLM Inference Efficiency with Heterogeneous Sliding-Window Lengths}

\author{%
Tianyu Fu$^{1,2,}$\thanks{Equal contribution. Genghan Zhang made contributions while at Tsinghua University},
Haofeng Huang$^{1,2,*}$,
Xuefei Ning$^{1,*}$,
Genghan Zhang$^{3}$,
Boju Chen$^{1}$,\\
\textbf{Tianqi Wu$^{1,2}$,
Hongyi Wang$^{1,2}$,
Zixiao Huang$^{1,2}$,
Shiyao Li$^{1,2}$, 
Shengen Yan$^{1,2}$,}\\
\textbf{Guohao Dai$^{2,4}$,
Huazhong Yang$^{1}$,
Yu Wang$^{1}$}
\\
\\
$^{1}$Tsinghua University $^{2}$Infinigence-AI $^{3}$Stanford University
$^{4}$Shanghai Jiao Tong University \\
}

\begin{document}

\ifcolmsubmission
\linenumbers
\fi

\maketitle



\begin{abstract}
\input{content/abstract}
\end{abstract}

\input{content/introduction}
\input{content/related_work}
\input{content/oracle}
\input{content/method}
\input{content/dataset}
\input{content/experiment}

\input{content/conclusion}

\clearpage
\input{content/aknowledge}

\bibliographystyle{colm2025_conference}
\bibliography{reference}

\clearpage
\appendix
\input{content/appendix}

\clearpage

\end{document}

%% file: macros.tex
\usepackage{xspace}
\usepackage{graphicx}
\usepackage{amsmath}
\usepackage{amssymb}
\usepackage{amsthm}
\usepackage{adjustbox}
\usepackage{wrapfig}
\usepackage{lipsum}
\usepackage{booktabs}

\newcommand{\name}[0]{MoA\xspace}
\newcommand{\xhdr}[1]{{\noindent\bfseries #1}.}
\newcommand{\rom}[1]{\textup{\uppercase\expandafter{\romannumeral#1}}}

\usepackage{pifont}
\newcommand{\cmark}{\ding{51}}%
\newcommand{\xmark}{\ding{55}}%

\usepackage[most]{tcolorbox}
\usepackage{colortbl}
\usepackage{pifont}
\usepackage{makecell}

\newtcolorbox[
  auto counter,
  list inside=examplelist,
]{greenbox}[2][]{
  enhanced,
  title={Example~\thetcbcounter. \textbf{#1}},
  colback=green!5!white,
  colframe=green!50!black,
  colbacktitle=green!60!black,
  coltitle=white,
  #2
}

\newtcolorbox[
  auto counter,
  list inside=examplelist
]{bluebox}[2][]{
  enhanced,
  title={Format~\thetcbcounter. \textbf{#1}},
  colback=cyan!5!white,  
  colframe=cyan!50!black,  
  colbacktitle=cyan!75!black,  
  coltitle=white,
  #2
}



%% file: content/abstract.tex
Sliding-window attention offers a hardware-efficient solution to the memory and throughput challenges of Large Language Models (LLMs) in long-context scenarios.
Existing methods typically employ a single window length across all attention heads and input sizes. However, this uniform approach fails to capture the heterogeneous attention patterns inherent in LLMs, ignoring their distinct accuracy-latency trade-offs. 
To address this challenge, we propose \textit{Mixture of Attention Spans} (\name), which automatically tailors distinct sliding-window length configurations to different heads and layers. \name constructs and navigates a search space of various window lengths and their scaling rules relative to input sizes. It profiles the model, evaluates potential configurations, and pinpoints the optimal length configurations for each head. 
\name adapts to varying input sizes, revealing that some attention heads expand their focus to accommodate longer inputs, while other heads consistently concentrate on fixed-length local contexts. Experiments show that \name increases the effective context length by 3.9$\times$ with the same average sliding-window length, boosting retrieval accuracy by 1.5-7.1$\times$ over the uniform-window baseline across Vicuna-\{7B,13B\}, and Llama3-\{8B,70B\} models. Moreover, \name narrows the performance gap with full attention, reducing the maximum relative performance drop from 9\%-36\% to within 5\% across three long-context understanding benchmarks. \name achieves a 1.2-1.4$\times$ GPU memory reduction, boosting decode throughput by 6.6-8.2$\times$ and 1.7-1.9$\times$ over FlashAttention2 and vLLM, with minimal performance impact. 
Our code is available \href{https://github.com/thu-nics/MoA}{here}.

%% file: content/introduction.tex
\section{Introduction}
\label{sec:introduction}

Large Language Models (LLMs) exhibit remarkable versatility across numerous applications~\citep{brown2020gpt3, tay2022efficientSurvey, Wan2023EfficientLLM}.
Central to LLM is the attention mechanism~\citep{vaswani2017attention}, which computes interactions among tokens within a certain span, thereby enabling context understanding.
Scaling input length is crucial for enhancing LLM capabilities~\citep{Chen2023PI, Tworkowski2023FocusedTrans}, including fact retrieval, summarization, few-shot learning, question answering and so on~\citep{bai2023longbench, yuan2024lveval}. 
However, the ever-growing attention computation and Key-Value Cache (KV-Cache) pose significant efficiency challenges~\citep{xiao2023streamingLLM}.


\begin{figure}[ht]
    \vspace{-15pt}
    \centering
    \includegraphics[width=\textwidth]{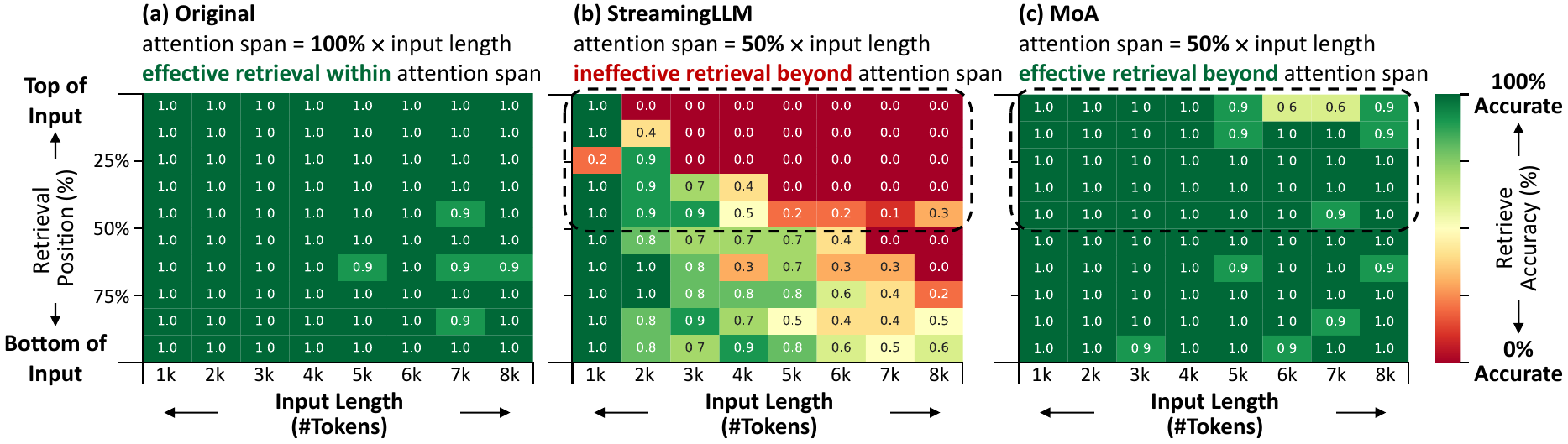}
    \vspace{-15pt}
    \caption{Retrieval accuracy of the Vicuna-7B model using different attention methods across varying input lengths and retrieval positions on the LongEval benchmark~\citep{lmsys2023longeval}. The benchmark takes massive key-value pairs as inputs and tests the accuracy to retrieve values based on given keys from diverse positions. (a) Original model with a full attention span; (b) StreamingLLM with half the attention span, showing reduced effectiveness beyond the span; (c) \name with half the attention span, maintaining effectiveness beyond the span.}
    \vspace{-4pt}
\label{fig:retrieve_acc_context_position}
\end{figure}

Previous work has proposed sliding-window attention to address the efficiency challenges of long contexts in generative LLMs. These methods typically employ a uniform, fixed-span sliding-window mask across all heads and input lengths, limiting attention to local span only~\citep{xiao2023streamingLLM, han2023lmInfinite}.
This design allows the LLM to process long inputs within a bounded attention computation and KV caching overhead.
Following previous works~\citep{Chen2023PI, Tworkowski2023FocusedTrans}, we quantify the effective context length as the maximum input length where content retrieval accuracy exceeds a 90\% threshold. 
In principle, fixed-span local attention can gradually aggregate global information through multiple model layers, yielding a longer effective context length than each attention span~\citep{feng2022diffuser, zaheer2020bigbird}.
Nonetheless, we reveal that uniform sliding-window methods, such as StreamingLLM~\citep{xiao2023streamingLLM}, hardly extend effective context length beyond the span.
As shown in Figure~\ref{fig:retrieve_acc_context_position}(b),
with a 50\% attention span, StreamingLLM fails to accurately retrieve earlier half of the input and performs even worse at longer input lengths.
Figure~\ref{fig:attention_oracle} reveals one possible explanation for the problem: while some attention heads focus on local contexts, others encompass the broad span of the entire input sequence. 
Consequently, the uniform approach fails to achieve a long effective context length, as it limits the attention span of global-context heads while excessively allocating compute and memory budgets to local-context heads. 
Additionally, as the input length increases, some attention heads need a faster increase in attention span than others to avoid serious performance degradation, as shown in Table~\ref{tab:oracle_layer_length}. Unfortunately, the uniform approaches do not include heterogeneous rules to scale the attention spans differently for various heads.
In addition, existing model compression methods, including quantization and sparse attention~\citep{smooth_quant, llm-mq, squeezellm, li2024evaluating}, typically rely on \textit{general language modeling corpora} to determine compression configurations, which fail to accurately capture their impact on \textit{long-context tasks}.

In this work, we propose Mixture of Attention Spans (\name), a training-free heterogeneous sliding-window attention method.
As shown in Figure~\ref{fig:overview}, \name constructs a search space of heterogeneous and elastic rules that scale each head’s window length with input length.
To automate rule selection, \name first utilizes gradient-based profiling to inspect the influences of each attention position on the prediction loss. 
Based on the profiling results, \name tailors heterogeneous window length for each attention head and input length.
During profiling, \name employs a calibration dataset with long-range summaries from a full-attention model, instead of the human-written summaries, as the reference to calculate the loss. This ensures an accurate profiling of attention influences to facilitate better performance.
Our contributions are summarized as follows.

\begin{itemize}
    \item \textbf{Heterogeneous Elastic Rules}.
    We propose heterogeneous elastic rules for sliding-window length of each attention head.
    We formulate \name configuration search space to include a diverse range of elastic rules that tailor the local attention span relative to the input length for each attention head.
    The heterogeneous elastic rules improve content retrieval accuracy from 25\% to 98\% over the uniform counterpart.
    
    \item \textbf{Calibration Dataset Construction}
    We emphasize the importance of data engineering in LLM compression. Our findings demonstrate that, instead of relying on general language modeling datasets and human responses, using datasets with long-range dependencies and referencing the original LLM’s responses is essential for accurately profiling the effects of compression.
    
    \item \textbf{Automatic Configuration Search}.
    We propose an automatic pipeline to find the optimal configuration encompassing heterogeneous elastic rules for various attention heads. This pipeline efficiently finds the optimal configuration within several hours—for example, two hours for Vicuna-13B.
    
\end{itemize}


Experiments show that \name achieves 6.6-8.2$\times$ throughput improvements over dense FlashAttention2 on 7B and 13B LLMs at a 50\% density (average KV-Cache length / input length). The significant throughput improvements stem from four factors: (1) the static size of the KV-Cache, (2) reduced attention computations, (3) increased batch sizes enabled by reduced memory usage, and (4) a specialized kernel implementation. 
Moreover, \name matches the performance of full-attention models across various long-context retrieval and understanding benchmarks, exhibiting an average relative performance drop below 1\%, roughly one-fifth that of uniform sliding-window attention baselines. 
Even at just 25\% density, \name achieves over 90\% retrieval accuracy, significantly outperforming baselines that require 75\%–100\% density for similar performance.
Our code is available at \url{https://github.com/thu-nics/MoA}.

%% file: content/related_work.tex
\section{Preliminary and Related Work}
\subsection{Attention Mechanism}

The Multi-Head Self-Attention (MHA) mechanism~\citep{vaswani2017attention} is crucial to the functionality of LLMs. It starts with an input sequence transformed into query (Q), key (K), and value (V) matrices through linear projections. These matrices, combined with the cached K and V (KV-Cache) from previous sequences, compute the attention matrix (A). This calculation is modified by a causal mask (M) to ensure autoregressive properties, resulting in the output (O), as depicted in Equation~\ref{eq:attention_compute}:
\begin{equation}
\begin{aligned}
\mathbf{S} = \mathbf{Q} \mathbf{K}^T, \quad \mathbf{A} = \text{softmax}(\mathbf{S}+\mathbf{M}), \quad \mathbf{O} = \mathbf{A} \mathbf{V}
\end{aligned}
\label{eq:attention_compute}
\end{equation}
Autoregressive inference in LLMs involves two stages: prefill and decode. 
During prefill, the model processes the entire input sequence to generate the initial response token. 
In the subsequent decode stage, it iteratively uses the newly generated token and previously cached K, V matrices to produce subsequent tokens. 
Although effective, this iterative process increases memory and computation demands due to the expanding KV-Cache. 

\subsection{Efficient Attention Methods}
\label{sec:related_work}

Efficient attention methods have been proposed to mitigate the computation and memory costs. 
One branch of work \textbf{dynamically} skip attention computations during the prefill stage~\citep{Pagliardini2023SparseFlash, qu2022dota, roy2021routingTransformer, wang2021spatten, lu2021sanger, kitaev2020reformer} or drop the KV-Cache during the decode stage~\citep{Anagnostidis2023DynamicPruning, Zhang2023H2O, Ge2023FastGen,sheng2023flexgen, Liu2023Scissorhands} based on the input sequence. 
However, due to the complex control and computation flow, dynamic sparse prefill often requires specific hardware to achieve substantial wall-time speedups~\citep{qu2022dota, wang2021spatten, lu2021sanger, ham2021elsa, ham2020A3}. 
Additionally, dynamic KV-Cache pruning in the decode stage may require extensive retraining~\citep{Anagnostidis2023DynamicPruning}, additional pruning score computation~\citep{sheng2023flexgen, Zhang2023H2O, Liu2023Scissorhands, Ge2023FastGen, li2024snapkv, cai2024pyramidkv}, or extensive memory swapping for KV-Cache retrieval~\citep{tang2024quest, xiao2024infllm}.

Another branch of work uses \textbf{static} sparse attention, where predefined masks are applied consistently across all input sequences.
Thanks to the fixed computation flow, static sparse attention is generally more efficient and GPU-friendly.
For language understanding models such as BERT~\citep{devlin2018bert}, various masks are used~\citep{zaheer2020bigbird, beltagy2020longformer, child2019spTrans, Zhou2024EfficientSurvey,xiao2023streamingLLM, han2023lmInfinite}. 
But for generative LLMs, the predominant method is the uniform sliding-window mask with global attention on a few initial tokens~\citep{xiao2023streamingLLM, han2023lmInfinite}.
For these approaches, the KV-Cache beyond the local attention span can be dropped, saving much memory for long sequence scenarios.
However, uniform sliding-window masks across different attention heads and input lengths are model- and data-agnostic, which can compromise LLMs' effective context length and lead to suboptimal performance in long-sequence scenarios.
Our method falls within this category, benefiting from the efficiency and training-free advantages, while addressing the performance limitations encountered by previous methods.



Previous works also propose LLM acceleration frameworks~\citep{huggingface2022accelerate,Aminabadi2022DeepSpeed,sheng2023flexgen,Kwon2023vllm}, as well as kernel-level optimizations~\citep{dao2022flashattention, dao2023flashattention2, shah2024flashattention3}.
These kernel and system optimizations are orthogonal to our work and can be integrated to further enhance efficiency.



%% file: content/oracle.tex
\section{Mixture of Attention Spans (\name)}
\label{sec:mixture_of_attention}

We first illustrate the heterogeneity of attention patterns in pre-trained LLMs in Section~\ref{sec:elastic_attention}. 
Based on this insight, we define the search space for our \name method in Section~\ref{sec:mask_search_space}.

\begin{figure}[!t]
\centering
\vspace{-15pt}
\begin{minipage}[b]{0.55\textwidth}
    \centering
    \includegraphics[width=\textwidth]{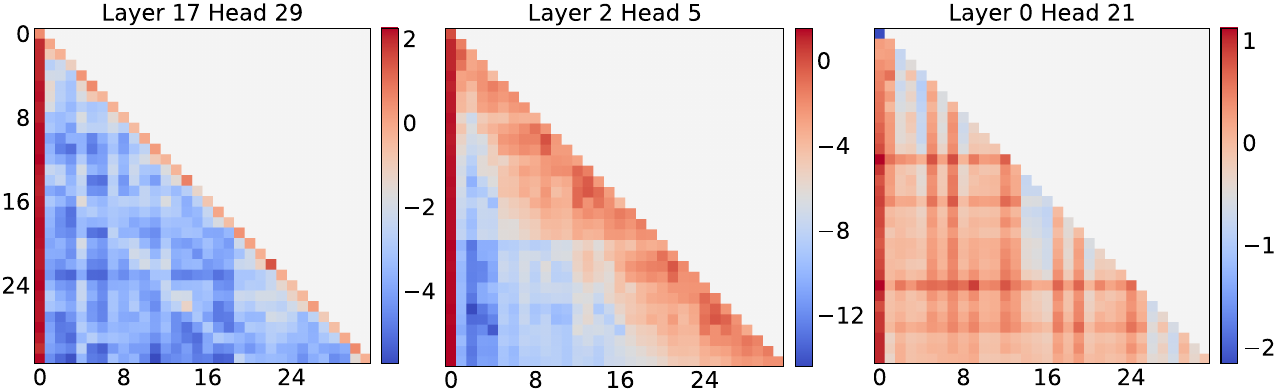}
    \caption{Examples of attention matrices from different attention heads of the Vicuna-7B model. Each attention matrix is averaged over 256 data items from the LongEval dataset.}
    \label{fig:attention_oracle}
\end{minipage}
\hfill
\begin{minipage}[b]{0.40\textwidth}
    \centering
    \setlength\tabcolsep{4pt}
    \begin{tabular}{l|c|c|c}
    \toprule
        ~ & \multicolumn{3}{c}{Window/Input Len.} \\ 
        \multirow{1}{*}{Layers} & 2k/4k & 2k/8k & 4k/8k\\
        \midrule
        6,\space\space7,\space\space8 & \color{Maroon} 0.83 & \color{Maroon} 0.29 & \color{Maroon} 0.61\\ 
        9,\space10,11 & \color{OliveGreen} \textbf{0.99} & \color{Maroon} 0.81 & \color{OliveGreen} 0.96\\ 
        17,18,19 & \color{OliveGreen} 0.97 & \color{OliveGreen} \textbf{0.94} & \color{OliveGreen} \textbf{0.97}\\ 
    \bottomrule
    \end{tabular}
    \captionof{table}{Retrieval accuracy of Vicuna-7B with sliding-window attention across various model layers, window spans, and input lengths.}
    \label{tab:oracle_layer_length}
\end{minipage}
\vspace{-4pt}
\end{figure}

\subsection{Mixture of Attention Patterns and Elastic Rules}
\label{sec:heterogeneous_attention}
\label{sec:elastic_attention}

\xhdr{Heterogeneous Attention Patterns} Different attention heads in LLMs exhibit heterogeneous attention patterns, as shown in Figure~\ref{fig:attention_oracle}. 
For example, the first head primarily focuses on local contexts with a narrow-span sliding window, while the third head covers nearly the entire input, indicating global attention.
The attention spans of different heads mostly remain constant across various tasks and datasets, as shown in Appendix~\ref{sec:appendix/oracle}.
Table~\ref{tab:oracle_layer_length} demonstrates that applying the same sliding-window attention mask across model layers can lead to a 65\% variance in retrieval accuracies. 
This aligns with the multi-head self-attention design principle of capturing varied information~\citep{vaswani2017attention}, as well as findings from concurrent research that identify specific attention heads for global text retrieval~\citep{wu2024retrievalHead}.

\xhdr{Heterogeneous Elastic Rules} In addition to heterogeneity at a certain length, different attention heads also exhibit varying elastic behaviors as the input length changes. 
Figure~\ref{fig:attention_oracle} illustrates this variability: 
for shorter inputs (the upper-left part of the attention matrix), the second and third heads initially show global attention. However, as input length increases, the second head maintains a medium-span local focus, while the third head continues to expand as global attention.
Table~\ref{tab:oracle_layer_length} further evidences the diverse elastic rules. For example, at a 4k input length, a 2k sliding-window attention mask on layers 9 to 11 yields better retrieval accuracy than on layers 17 to 19; however, the opposite holds for an 8k input length. These data support the visual observations from Figure~\ref{fig:attention_oracle}, highlighting that attention patterns respond differently to input-length scaling.
Leveraging these insights, \name encompasses heterogeneous elastic rules as the search space.

\subsection{Heterogeneous Elastic Rule Search Space}
\label{sec:mask_search_space}

In designing the search space for \name, we consider the inherently heterogeneous and elastic nature of LLM attention patterns.
As shown in Figure~\ref{fig:overview}(a), we adopt a hardware-friendly sliding-window mask as our base attention mask~\citep{beltagy2020longformer}. Following previous work~\citep{xiao2023streamingLLM, han2023lmInfinite}, the initial few tokens (64 tokens for \name) are not masked. The attention span equals the sliding-window span plus the number of initially unmasked tokens.
We define the attention span $\textit{S}$ of head $h$ at input length $\textit{N}$ using a straightforward linear function:
\begin{equation}
    S_h = \alpha_h + \beta_h \cdot N,
    \label{eq:search_space}
\end{equation}
where $\alpha_h$ and $\beta_h$ are hyperparameters that control the base span and its expansion rate with input length for a specific attention head. 

The $\alpha$ and $\beta$ hyperparameters for each attention head are chosen from multiple discrete options. By default, \name uses six and nine options for $\alpha$ and $\beta$, respectively. For LLMs with many heads and layers, the search space can become quite large. 
For example, for a 7B model consisting of $32$ attention heads and $32$ layers, the potential search space expands to $54^{1024}$ configurations. Thus, we design an automatic pipeline to efficiently pinpoint the optimal $\alpha$s and $\beta$s for any LLM.

%% file: content/method.tex
\section{Automatic \name Configuration Search Pipeline}
\label{sec:automatic_pipeline}
\begin{figure}[tb]
\vspace{-15pt}
\centering
    \includegraphics[width=0.9\textwidth]{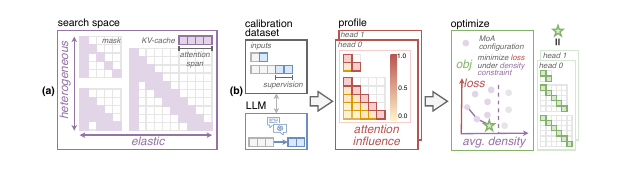}
    \caption{
    Overview of \name. 
    (a) The search space includes heterogeneous elastic rules of the attention span on sliding-window masks.
    (b) The automatic rule search pipeline begins with a calibration dataset, which includes long-dependency contexts and supervision texts generated by the original dense LLM. 
    \name profiles each attention value's impact on model predictions within this dataset, revealing accuracy losses for different candidate elastic rules across various input lengths. 
    The final optimization step selects elastic rules for each attention head to minimize the total prediction loss while adhering to specified density constraints.
    }
    \label{fig:overview}
    \vspace{-4pt}
\end{figure}

This section outlines the \name automatic configuration search pipeline, as shown in Figure~\ref{fig:overview}(b). 
Starting with a trained LLM and a calibration dataset, \name first \textbf{profiles} the influence of each attention value on the model's prediction loss for various input sequences from the calibration dataset.
The masked sum of the influences represents the accuracy loss associated with each mask at different input lengths, showing the accuracy loss each candidate elastic rule could cause at that length.
Then, \name \textbf{optimizes} the window-lengths by selecting the optimal elastic rule for each head, which minimizes the accuracy loss across various input lengths while adhering to specified density constraints.
The following sections provide detailed discussions of each step in this pipeline. 
Detailed definitions of key notions are in Appendix~\ref{sec:appendix/notions}.
The user-friendly code interface is described in Appendix~\ref{sec:appendix/interface}.

\subsection{Attention Influence Profiling}
\label{sec:pipeline/profile}

In the profiling step, \name quantifies the impact of individual attention values on the final prediction loss of a pre-trained LLM. 
It informs the subsequent step about the influence of masking each attention value, revealing the accuracy trade-offs of the candidate elastic rules for each attention head.

The influence of each attention value is derived from the attention matrix \(\textbf{A}\) and its gradient \(\partial L / \partial \textbf{A}\), computed over a calibration dataset. 
When applying sliding-window attention masks, we approximate the change in the model's prediction loss, \(\Delta L\), using a first-order Taylor expansion based on variations in the attention matrices \(\textbf{A}\): $\Delta L = \sum_h \sum_i \sum_j \partial L/\partial A_{h,i,j} \cdot \Delta A_{h,i,j}$.
Here, \(h\) indexes the attention heads across all layers, and \(i, j\) are the row and column indices within each attention matrix \(\textbf{A}_{h}\).
Details on the calibration dataset and the prediction loss $L$ are provided in Section~\ref{sec:dataset}.

We define the \textit{attention influence} matrix, \(\textit{E}_{h,i,j}\), as the estimated change in loss, \(\Delta L\), if the attention value \(A_{h,i,j}\) is masked (i.e., set to zero).
As shown in Equation~\ref{eq:effect}, this measure considers both the direct and indirect effects of the mask. For notation simplicity, we omit the head index $h$ here.
Initially, masking directly reduces the attention value to zero, represented by \(\Delta A_{i,j|j} = -A_{i,j}\). 
Additionally, the \(\text{softmax}\) function in attention normalizes the sum of each row in the attention matrix to one. Thus, setting one attention value at column $j$ to zero causes an increase in the other attention values, $\Delta A_{i,n|j}, n\neq j$, within the same row. 
These two effects are integrated into the following formulation, whose derivation is provided in Appendix~\ref{sec:appendix/derive_attention_influence}:
\begin{equation}
    E_{i,j} = \sum_n \frac{\partial L}{\partial A_{i,n}} \cdot \Delta A_{i,n|j} 
    = \frac{\partial L}{\partial A_{i,j}} \cdot (-A_{i,j}) 
    + \sum_{n\neq j} \frac{\partial L}{\partial A_{i,n}} \cdot A_{i,n} \cdot \frac{A_{i,j}}{1 - A_{i,j}}
\label{eq:effect}
\end{equation}
In practice, we use backpropagation on a calibration dataset to calculate the average attention influence \(\bar{\textbf{E}}_h\) of each head across data items. The average attention influence is calculated respectively for different input lengths.
The gradient \({\partial L}/{\partial \textbf{A}_h}\) is computed using chain derivative rule in deep learning frameworks like PyTorch~\citep{paszke2019pytorch}. The detailed calibration dataset setup is discussed in Section~\ref{sec:dataset}.

With the average attention influence of each head, \name can calculate the accuracy loss of applying a candidate elastic rule at a specific input length.
The loss is calculated as the sum of masked average attention influence according to the rule.
We denote \(\textbf{M}_{r_h}\) as the binary mask at head $h$ that corresponds to rule $r$, with masked positions marked as 1 and others as 0. We formalize accuracy loss $\Delta L$ as follows:
\begin{equation}
    \Delta L = \sum_h \Delta L_{h,r_h} = \sum_h \sum_i \sum_j M_{r_h, i, j} \cdot \bar{E}_{h, i, j}.
    \label{eq:assess}
\end{equation}

After the profiling stage, \name acquires the estimated accuracy impact of elastic rules across attention heads. 
This informs the allocation of longer windows to more sensitive heads and shorter windows to less sensitive ones.
Profiling at different input lengths enables the identification of the most effective elastic rules, even for unseen lengths.

\subsection{Automatic Optimization}
\label{sec:pipeline/optimize}

\name automatically selects the optimal elastic rule for each attention head to minimize accuracy losses across various sequence lengths under density budgets. 
Based on the profiling results, \name first identifies Pareto-front configurations where any improvement in accuracy loss at one profiled length would worsen another. 
To ensure the best generalization to lengths beyond those profiled, \name then selects the configuration that yields the minimum loss at an unseen length among the Pareto front solutions as the final configuration.

Specifically, we utilize multi-objective optimization to search for a set of Pareto optimal configurations across the profiled lengths. 
The objective for each length is to minimize the total accuracy loss while conforming to any user-defined density constraints.
The objective is formulated as follows: 
\begin{equation}
    \mathop{\arg\min}_{r_h \in \mathbb{R}} \Delta L^{(N_i)}, N_i \in \mathbb{N}_{\text{profile}} 
    \quad \operatorname{s.t.} \space
    \frac{1}{H}\sum_{h=1}^H d^{(N_i)}_{r_h} \leq d^{(N_i)}_\text{constr}, \forall N_i \in \mathbb{N}_{\text{constr}}.
    \label{eq:multi-optimize}
\end{equation}
Here, superscript \((N)\) denotes values at different lengths;
\(\mathbb{N}_{\text{profile}}\) and \(\mathbb{N}_{\text{constr}}\) denote the sets of lengths for profiling and those subject to density constraints, respectively;
$\mathbb{R}$ denotes the set of candidate rules;
\( \Delta L^{(N_i)} \) denotes the accuracy loss due to applying attention mask;
\( d_{r_h}^{(N_i)} \) denotes the density of rule $r_h$ at head $h$; 
\( d_{\text{constr}}^{(N_i)} \) denotes the average density constraint;
\(H\) denotes the total number of attention heads.

This formulation corresponds to the classic multi-objective mixed-integer-programming problem, which can be effectively solved within minutes using existing linear solvers, like Gurobi~\citep{gurobi}. The detailed formulation and solving strategies are discussed in Appendix~\ref{sec:appendix/optimization_details}.

Among the Pareto-optimal \name configurations, we select the one with the minimum loss at the unseen validation length as the optimal solution. 
This avoids profiling at every possible length and increases the likelihood of effective generalization to unseen input lengths.


%% file: content/dataset.tex
\section{Dataset and Supervision}
\label{sec:dataset}

\begin{table}[tb]
    \centering
    \footnotesize
    \vspace{-15pt}
    \setlength\tabcolsep{3pt}
    \begin{tabular}{@{}lll|cc|ccc@{}}
    \toprule
    Dataset & Supervision & Reference & Long Dep. & Align Model & Retrieval Acc. $\uparrow$ & PPL $\downarrow$ \\
    \midrule
    RedPajama & Context & - & \xmark & \xmark & 0.25 & 4.95 \\
    MultiNews & Context \& Summary & Human & \xmark /\cmark & \xmark & 0.27 & 4.62 \\
    MultiNews & Summary & Human & \cmark & \xmark & 0.87 & 3.97 \\
    MultiNews & Summary & Model & \cmark & \cmark & \textbf{0.95} & \textbf{3.96} \\
    \bottomrule
    \end{tabular}
    \vspace{-4pt}
    \caption{Calibration dataset design choices: dataset content, supervision, and response reference. Calibration dataset with long dependency and model alignment improves \name performance on retrieval accuracy and perplexity. All tests are done at 25\% average density at 8k input length.
    }
    \vspace{-4pt}
    \label{tab:dataset}
\end{table}

We emphasize the often-overlooked role of calibration dataset design and supervision objectives in LLM compression.
Calibration datasets enable effective sensitivity analysis across compression methods like weight pruning~\citep{Men2024ShortGPT,lee2024cats} and quantization~\citep{awq,smooth_quant}.
In this work, \name profiles the attention influence on the calibration dataset, which is crucial for subsequent automatic optimization. 

\xhdr{Current Approach} 
General language modeling datasets, such as the human-written text corpus RedPajama~\citep{together2023redpajama}, are commonly used as calibration datasets. These datasets, supervised by next-token prediction on the entire corpus, primarily capture attention patterns coherent with immediately preceding tokens.
However, they lack long context dependencies, failing to address the global attention crucial for long-context tasks. 

Moreover, a notable misalignment exists between the model response and human-written supervision. Consequently, this leads to inaccuracies when using human responses to compute attention values and gradients during profiling.
For example, given the same question, a human might answer `Blue', while the model could generate `The blue color'.
When using the human answer for supervision, attention influence is inaccurately quantified based on the probability shift for predicting `Blue'; this diverges from the objective of maintaining crucial attention for the original model prediction, `The'.
These inconsistencies arise from various factors, including mismatched positions, tones, and synonyms.

\xhdr{\name's Approach} 
\name enhances the calibration dataset by integrating \textit{long-range dependencies} and \textit{model alignment}.
Specifically, we utilize the long-context MultiNews dataset~\citep{Fabbri2019MultiNews}, which includes summaries that depend heavily on long-range content. The summaries are generated by the original dense model and serve as supervision.
Compared to current approaches that adopt human responses as the reference to calculate the cross-entropy loss $L$, using the responses generated by the original model as the supervision facilitates accurate profiling, thus benefiting the \name configuration search.

\xhdr{Approach Comparison} 
We validate our design with varied dataset choices, supervision types, and summary references, while standardizing the data item count and length to 50 and 8k tokens, respectively. Detailed setups and evaluations are in Appendices~\ref{sec:appendix/experiment/experiment_setup} and~\ref{sec:appendix/experiemnt/ablation/calibration_dataset}.

We show the importance of long-range dependencies by comparing the \name configuration generated with different datasets and supervisory methods. 
In Table~\ref{tab:dataset}, RedPajama~\citep{together2023redpajama} represents the general language modeling dataset, while MultiNews~\citep{Fabbri2019MultiNews} highlights long-range contexts by aggregating multiple documents on a single incident.
MultiNews additionally provides human-written summaries, enhancing long-range dependencies.
Using MultiNews summaries for loss calculation significantly improves retrieval accuracy by 60\% and reduces perplexity by 0.98.

Furthermore, using summaries generated by the original dense model 
as supervision promotes higher alignment between its own attention patterns and the text supervision. It improves performance compared to potentially inconsistent human summaries, as shown in the last two rows of Table~\ref{tab:dataset}.

%% file: content/experiment.tex
\section{Experiment}
\label{sec:experiment}

\begin{figure}[t]
    \vspace{-20pt}
    \centering
    \begin{minipage}[b]{0.53\textwidth}
        \centering
        \includegraphics[width=0.93\textwidth]{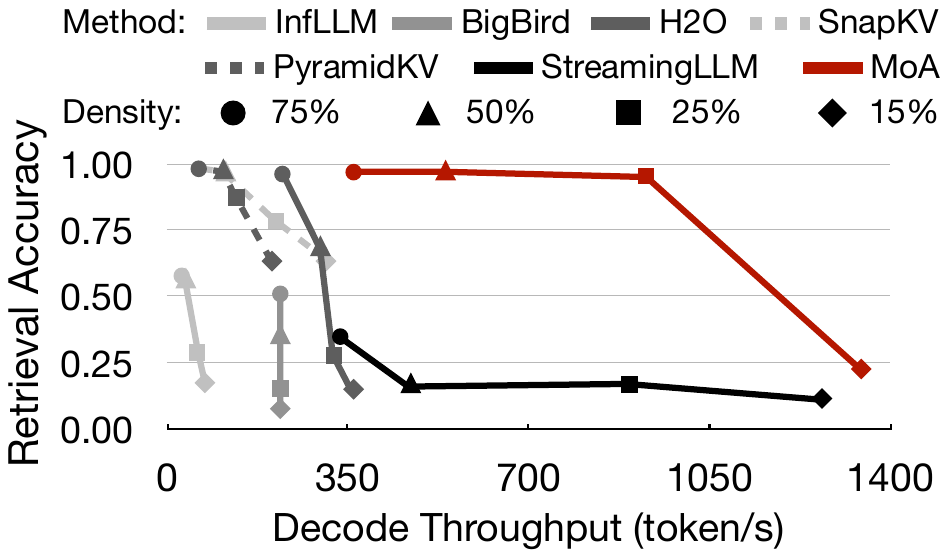}
        \vspace{-5pt}
        \caption{Accuracy-throughput trade-offs of seven attention methods at different densities, tested on Vicuna-7B with 8k input length using one A100-80GB GPU on the LongEval dataset.}
        \label{fig:pareto_front}
    \end{minipage}\hfill
    \begin{minipage}[b]{0.43\textwidth}
        \centering
        \footnotesize
        \setlength\tabcolsep{3pt}
        \begin{tabular}{l|cc|cc}
        \toprule
        ~ & \multicolumn{2}{c|}{Retrieval Acc.} & \multicolumn{2}{c}{PPL} \\
        Mask Design & 8k & 16k & 8k & 12k \\
        \midrule
        Uniform & 0.25 & 0.15 & 4.89 & 5.19 \\
        +Hetero. Layers & 0.31 & 0.26 & 4.55 & 4.85 \\
        +Hetero. Heads & 0.95 & 0.41 & \textbf{3.96} & 4.30 \\
        +Elastic & \textbf{0.98} & \textbf{0.43} & \textbf{3.96} & \textbf{4.29} \\
        \bottomrule
        \end{tabular}
        \captionof{table}{Ablation study on search space with consistent 25\% density, progressively introducing heterogeneity in layers, heads, and elastic rules. Evaluations are done with retrieval accuracy and perplexity.}
        \label{tab:mask_design}
    \end{minipage}
    \vspace{-12pt}
\end{figure}

\subsection{Setups}
\label{sec:experiment_setup}
We briefly summarize the experimental setup here; more details appear in Appendix~\ref{sec:appendix/experiment/experiment_setup}.

\xhdr{Baselines} We compare \name with state-of-the-art static and dynamic efficient attention methods, including StreamingLLM~\citep{xiao2023streamingLLM}, InfLLM~\citep{xiao2024infllm} and H2O~\citep{Zhang2023H2O}.
We define the \textit{density} as the ratio of the average in-memory KV-Cache length to sequence length during decoding.
Notably, \name and StreamingLLM use efficient prefill, while H2O and InfLLM use the original dense prefill, with additional computations to dynamically determine the KV-Cache for decoding.

\xhdr{Models and Benchmarks}
We use vicuna-\{7b, 13b\}-v1.5-16k models~\citep{vicuna2023} from LMSys and Llama-3-\{8b, 70b\}-Instruct-262k models~\citep{meta2024llama3} from Gradient AI. 
We test long-context retrieval, understanding, and coherence abilities, using LongEval~\citep{lmsys2023longeval} retrieval accuracy, LV-Eval~\citep{yuan2024lveval}, and LongBench~\citep{bai2023longbench} scores, as well as average perplexity across four datasets~\citep{Dasigi2021Qasper,Fabbri2019MultiNews,li-roth-2002-trec1,hovy-etal-2001-trec2,lcc}. 
Performance experiments are restricted to eight A100-80GB GPUs over a 24-hour period, with Out-Of-Memory (OOM) and Out-Of-Time (OOT) conditions noted. 

\xhdr{\name Settings}
We restrict the number of distinct rules to at most two per model layer to ensure inference-time efficiency.
We profile \name on MultiNews~\citep{Fabbri2019MultiNews} with model summaries at 2k, 4k, and 8k lengths. The optimal \name configuration is selected using the validation dataset at 12k. Each model uses the same plan across all benchmarks and lengths. The models are not fine-tuned.

\subsection{Accuracy-Throughput Trade-off}

Figure~\ref{fig:pareto_front} demonstrates that \name advances the Pareto front in retrieval accuracy and decode throughput compared to six baselines.
At the same densities, \name notably improves throughput by 1.6-18.1$\times$ over H2O, InfLLM, BigBird~\citep{zaheer2020bigbird}, SnapKV~\citep{li2024snapkv}, and PyramidKV~\citep{cai2024pyramidkv}, driven by its efficient static attention design and customized GPU kernel. 
\name also maintains high retrieval accuracy across varying densities. Further evaluations across context lengths (4k-256k), model sizes (7B-70B), and benchmarks appear in subsequent sections.


\subsection{Performance}

\begin{table}[tb]
    \vspace{-20pt}
    \centering
    \vspace{-5pt}
    \setlength\tabcolsep{5pt}
    \begin{tabular}{ll|ccc|c|c|c}
    \toprule
    &  & \multicolumn{3}{c|}{Retrieve Acc. $\uparrow$} & LV-Eval $\uparrow$ & LongBench $\uparrow$& PPL $\downarrow$ \\
    Model & Attention  & 4k & 8k & 16k & 16k & 0-16k& 8-12k \\
    \midrule
    \multirow{5}{*}{Vicuna-7B} & Original  & 1.00 & 0.98 & 0.62 & 5.93 & 34.76& 3.79 \\
    \cmidrule(lr){2-8}
    & H2O  & 0.86 & 0.68 & 0.35 & 5.42 & 33.59& 3.94 \\
    & InfLLM  & 0.67 & 0.57 & 0.26 & 5.13 & 32.97& 4.07\\
    & StreamingLLM  & 0.43 & 0.16 & 0.08 & 4.72 & 31.84& 4.48 \\
    & \name  & \textbf{1.00} & \textbf{0.97} & \textbf{0.57} & \textbf{5.61} & \textbf{33.96} & \textbf{3.75} \\
    \midrule
    \multirow{5}{*}{Vicuna-13B} & Original  & 0.99 & 0.98 & 0.44 & 5.83 & 39.23& 3.62 \\
    \cmidrule(lr){2-8}
    & H2O  & 0.88 & 0.76 & 0.28 & 5.66 & 38.13& 3.80 \\
    & InfLLM  & 0.70 & 0.53 & 0.27 & 6.80 & 37.13& 4.07 \\
    & StreamingLLM  & 0.65 & 0.49 & 0.33 & 5.43 & 32.13& 4.10 \\
    & \name  & \textbf{0.99} & \textbf{0.93} & \textbf{0.49} & \textbf{7.16} & \textbf{38.77} & \textbf{3.62} \\
    \midrule
    \multirow{5}{*}{Llama3-8B} & Original  & 0.99 & 0.99 & 0.97 & 17.49 & 43.69& 4.52 \\
    \cmidrule(lr){2-8}
    & H2O  & 0.94 & 0.89 & 0.88 & 16.03 & \textbf{42.99} & 4.63 \\
    & InfLLM  & 0.65 & 0.59 & 0.37 & 14.44 & 42.43 & 4.68\\
    & StreamingLLM  & 0.68 & 0.55 & 0.52 & 11.16 & 38.22 & 4.79 \\
    & \name  & \textbf{0.99} & \textbf{1.00} & \textbf{1.00} & \textbf{17.46} & 42.97 & \textbf{4.49} \\
    \midrule
    \multirow{4}{*}{Llama3-70B} & Original  & 1.00 & 0.99 & 0.93 & 24.51 & 49.10& 3.67\\
    \cmidrule(lr){2-8}
    & H2O  & 0.93 & 0.91 & OOM & OOM & OOM & OOM \\
    & StreamingLLM  & 0.20 & 0.15 & 0.04 & 17.45 &  42.53 & 4.26\\
    & \name  & \textbf{1.00} & \textbf{1.00} & \textbf{0.94} & \textbf{23.65} & \textbf{47.79} & \textbf{3.75} \\
    \bottomrule
    \end{tabular}
    \caption{Comparative analysis of retrieval accuracy, LV-Eval scores, LongBench scores, and perplexity for various models with different attention methods. All methods employ 50\% density in decode stage. H2O uses dense prefill; StreamingLLM, InfLLM and \name use sparse prefill. InfLLM for 70B model is excluded due to OOT issues.
    }
    \label{tab:model_comparison}
    \vspace{-15pt}
\end{table}

\name outperforms state-of-the-art efficient attention methods across various model sizes and benchmarks, achieving comparable performance to full-attention model at 50\% density.

\begin{figure}[t]
\centering
\vspace{-25pt}
\begin{subfigure}[b]{0.38\textwidth}
    \centering
    \includegraphics[width=0.95\textwidth]{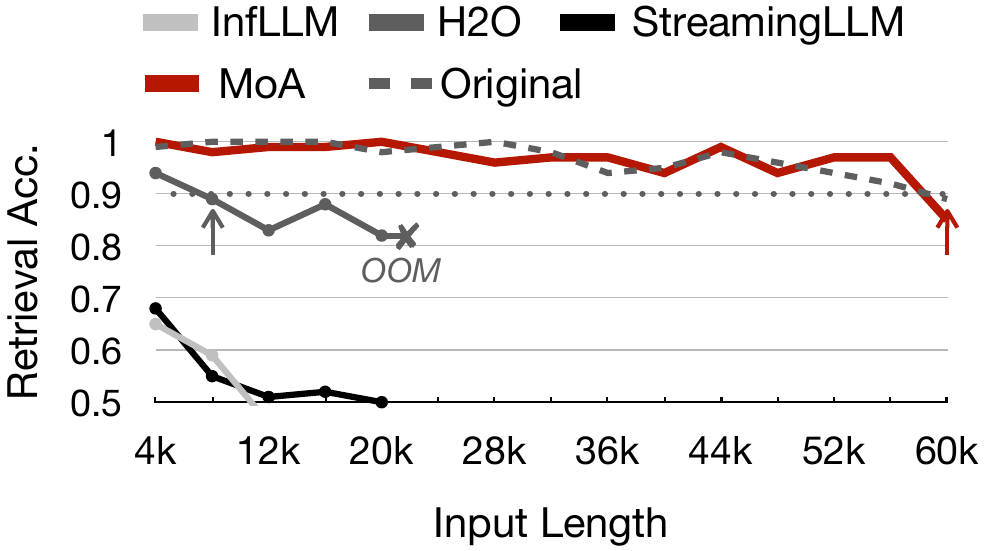}
    \vspace{-5pt}
    \caption{Retrieval accuracy and the effective context length (arrow).}
    \label{fig:effective_length}
\end{subfigure}
\hfill
\begin{subfigure}[b]{0.58\textwidth}
    \centering
    \footnotesize
    \setlength\tabcolsep{2pt}
    \begin{tabular}{l|cccc|ccc}
    \toprule
    ~  & \multicolumn{4}{c|}{Retrieve Acc. $\uparrow$} & \multicolumn{3}{c}{LV-Eval $\uparrow$}\\
    Attention  & 32k & 64k & 128k & 256k & 32k & 64k & 128k \\
    \midrule
    Original  & 0.98 & 0.93 & 0.76 & 0.37 & 16.74 & 15.39 & 14.71 \\
    \midrule
    InfLLM  & 0.43 & 0.32 & 0.25 & OOT & 14.22 & 12.17 & OOT \\
    StreamingLLM  & 0.52 & 0.48 & 0.41 & 0.25 & 12.38 & 11.45 & 11.94 \\
    \name  & \textbf{1.00} & \textbf{0.92} & \textbf{0.83} & \textbf{0.46} & \textbf{17.07} & \textbf{15.13} & \textbf{14.14} \\
    \bottomrule
    \end{tabular}
    \caption{Retrieval accuracy and LV-Eval score at longer lengths}
    \label{tab:scaling_context}
\end{subfigure}
\vspace{-4pt}
\caption{Experiments at extended sequence lengths with different attention methods using Llama3-8B model. All methods employ 50\% density in both prefill and decode stages.}
\label{fig:experiment/long_long_context}
\end{figure}

\xhdr{Long-Context Retrieval}
As shown in Table~\ref{tab:model_comparison}, \name demonstrates a maximum of 8\% relative accuracy drop (calculated as $\max\{1-\text{Acc.}_{\text{\name}} / \text{Acc.}_{\text{Original}}\}$ across three lengths and LLMs), significantly lower than StreamingLLM (87\%), InfLLM (58\%), and H2O (44\%). On average, \name’s relative accuracy drop is below 1\%, substantially better than the other methods (51\%, 41\%, and 20\%).
Figure~\ref{fig:experiment/long_long_context}(a) shows that \name retains over 90\% retrieval accuracy up to 60k lengths, equaling the dense model’s effective context length. Note that it is done within 8k profiling and 12k validation. In contrast, the effective context lengths for H2O, InfLLM, and StreamingLLM are only 8k, <4k, and <4k, respectively. 
Appendix~\ref{sec:appendix/experiment/retrieval} shows that \name extends its effective context to approximately $3.9\times$ the average KV-Cache length.

\xhdr{Long-Context Understanding}
Table~\ref{tab:model_comparison} indicates that \name reduces the maximum relative performance drop in LV-Eval and LongBench benchmarks to 5\% and 3\%, significantly less than StreamingLLM’s 36\% and 18\%. H2O and InfLLM incur maximum relative drops of 9\%-17\% and 3\%-5\%, respectively, but with greater efficiency costs. Figure~\ref{fig:lveval_longbench} and Table~\ref{tab:LVEval_breakdown} further show \name's consistent performance across subtasks, in contrast to the inconsistent performance of the baselines.
In perplexity tests, \name maintains less than a 1\% increase in relative perplexity, whereas other methods see increases of 4\%-13\%. These trends persist at other densities (Appendices~\ref{sec:appendix/experiment/overall_performance}, \ref{sec:appendix/experiment/long_context_understanding}). 

\xhdr{Longer-Context Generalization}
By compressing within 12k lengths, \name generalizes to lengths of 32k-256k, as shown in Figure~\ref{fig:experiment/long_long_context}(b). At the extended lengths, \name outperforms both InfLLM and StreamingLLM by 1.9-3.3$\times$ in retrieval accuracy and 1.2-1.4$\times$ in LV-Eval scores, demonstrating comparable performance to the original dense model. The configurations discovered by \name are detailed in Section~\ref{sec:insight}, supporting \name's strong generalizability across lengths.

\xhdr{Ablation Study}
We evaluate the performance impact of different sliding-window attention search spaces in Table~\ref{tab:mask_design}. 
Starting with a basic uniform sliding-window method, we observe significant enhancements by sequentially introducing heterogeneity: layers first, then heads, and finally elastic rules.

\subsection{Efficiency}
\label{sec:efficiency_test}

\begin{table}[tb]
    \footnotesize
    \centering
    \setlength\tabcolsep{1pt} 
    \vspace{-1pt}
    \begin{tabular}{@{}lll|cc|cc|cc}
    \toprule
    &  && \multicolumn{2}{c|}{4k} & \multicolumn{2}{c|}{8k} & \multicolumn{2}{c}{16k} \\
    Model &  Framework & Attention & Batch & Throughput & Batch & Throughput & Batch & Throughput \\
    \midrule
    \multirow{9}{*}{7B} &  vLLM & PagedAttention & 30 & 628.8 & 15 & 323.0 & 8 & 145.5 \\
        &  FlexGen & H2O & 20 & 754.9 & 6 & 296.3 & 1 & 51.7 \\
        &  HuggingFace & InfLLM & 15 & 62.0 & 10 & 37.5 & 6 & 19.2 \\
        &  HuggingFace & StreamingLLM & 50 & 945.1 & 25 & 467.3 & 12 & 232.0 \\
        \cmidrule(lr){2-9}
        &  \multirow{5}{*}{HuggingFace} & FlashAttention2 & 30 & 134.6 & 15 & 66.9 & 8 & 32.9 \\
        &  & +Static KV-Cache & 30 & 496.1 & 15 & 219.5 & 8 & 91.6 \\
        &  & +Reduced Attention & 30 & 722.5 & 15 & 369.9 & 8 & 178.3 \\
        &  & +Increased Batch & 50 & 897.7 & 25 & 436.7 & 12 & 206.4 \\
        &  & +Kernel (=\textbf{\name}) & 50 & \textbf{1099.0} & 25 & \textbf{535.7} & 12 & \textbf{257.3} \\
    \midrule
    \multirow{9}{*}{13B} &  vLLM & PagedAttention & 16 & 314.8 & 8 & 160.5 & 4 & 71.1 \\
        &  FlexGen &H2O & 12 & 330.2 & 4 & 138.2 & 1 & 37.4 \\
        &  HuggingFace & InfLLM & 8 & 30.3 & 5 & 17.63 & 3 & 11.3 \\
        &  HuggingFace & StreamingLLM & 28 & 478.4 & 14 & 241.2 & 7 & 116.5 \\
        \cmidrule(lr){2-9}
        &  \multirow{5}{*}{HuggingFace} &FlashAttention2 & 16 & 81.3 & 8 & 40.8 & 4 & 19.8 \\
        &  &+Static KV-Cache & 16 & 264.6 & 8 & 111.3 & 4 & 62.2 \\
        &  &+Reduced Attention & 16 & 329.6 & 8 & 156.4 & 4 & 87.3 \\
        &  &+Increased Batch & 28 & \text{471.5} & 14 & \text{222.6} & 7 & \text{108.3} \\
        &  &+Kernel (=\textbf{\name}) & 28 & \textbf{550.9} & 14 & \textbf{267.6} & 7 & \textbf{132.3} \\
    \bottomrule
    \end{tabular}
    \caption{Decode throughput (tokens/s) of Vicuna-{7B,13B}, evaluated at the maximum batch capacity of an A100-80GB GPU. All efficient attention methods use 50\% density. Efficiency improvements of \name are ablated with four factors. 
    }
    \vspace{-8pt}
    \label{tab:experiment/throughput}
\end{table}

\name shows high runtime efficiency with a manageable one-time search overhead.

\xhdr{Runtime}
Table~\ref{tab:experiment/throughput} compares \name's runtime efficiency across various attention methods and LLM frameworks, with an ablation of the efficiency improvements brought by each design factor of \name.
Implemented on HuggingFace, \name features a plug-and-play coding interface (Appendix~\ref{sec:appendix/interface}).
At 50\% density, \name achieves a decode throughput improvement of 6.6–8.2$\times$ over FlashAttention2 and surpasses H2O and InfLLM by 1.2–4.0$\times$.
Compared to the highly optimized vLLM framework~\citep{Kwon2023vllm}, \name still attains throughput gains of 1.7–1.9$\times$.
Additionally, \name reduces GPU memory usage by 1.2-1.4$\times$, as detailed in Appendix~\ref{sec:appendix/memory_throughput}. 
Results for 128k lengths are provided in Appendix~\ref{sec:appendix/128k_efficiency}.
This throughput gain results from four main factors: static KV-Cache size during generation ($\approx3.0\times$); reduced attention computations ($\approx1.5\times$); increased batch sizes from reduced KV-Cache memory ($\approx1.4\times$); and our CUDA GPU kernel for \name's heterogeneous attention ($\approx1.2\times$). 






\xhdr{Search Pipeline}
\name completes the automatic configuration search for the Vicuna-7B and 13B models within two hours. For the larger Llama3-70B model, the process requires 8.5 hours of wall time and 34.7 hours of GPU time. 
See Appendix~\ref{sec:appendix/pipeline_cost} for more details.


%% file: content/conclusion.tex
\section{Conclusion}
\label{sec:conclusion}
\name optimizes sliding-window lengths across attention heads by automatically searching heterogeneous elastic rules. 
With the same average window size, \name extends LLM context length by 3.9$\times$, significantly improving long-context retrieval and understanding over uniform-window baselines.
Additionally, it achieves throughput improvements of over $7\times$ compared to FlashAttention2, offering practical efficiency benefits.

%% file: content/aknowledge.tex
\section*{Acknowledgments}
This work was supported by the National Natural Science Foundation of China (No. 62325405, 62104128,
U19B2019, U21B2031, 61832007, 62204164), the Tsinghua EE Xilinx AI Research Fund, and the Beijing
National Research Center for Information Science and Technology (BNRist).
We thank Yi Ge for valuable discussions. 

%% file: content/appendix.tex
\section{Notations and Definitions}
\label{sec:appendix/notions}

To clarify the terminology and concepts introduced in our approach, we present the key notations and definitions used throughout the paper.

\subsection{Attention Span}

The \textbf{attention span} of each attention head is defined as the sum of a sliding-window length and a fixed prefix of always-visible tokens (attention sink). 
In the attention mask:

\begin{itemize}
    \item \textbf{Sliding Window:} A contiguous diagonal region indicating recent tokens each head can attend to.
    \item \textbf{Prefix:} The first 64 tokens of the input sequence, which remain visible to all attention heads regardless of the sliding-window length.
\end{itemize}

\subsection{Elastic Rule}

An \textbf{elastic rule} is a function $f(\cdot)$ parameterized by $\alpha$ and $\beta$, specifying how the attention span scales with the input sequence length $N$ for a single head. Formally, the attention span $\mathcal{S}$ given by an elastic rule is:
\begin{equation}
\mathcal{S}(N; \alpha, \beta) = \alpha \times N + \beta \label{eq:rule_to_span}
\end{equation}

Different attention heads employ different elastic rules by adopting different sets of $\alpha$ and $\beta$, resulting in heterogeneous attention spans optimized per head.

\subsection{\name Configuration}
An \textbf{\name configuration} is the collection of elastic rules of all attention heads of all layers in a model.

\subsection{Attention Mask}

\subsubsection{Visualization}

An \textbf{attention mask} is a binary mask applied to each head individually to gate its attention computation. It is defined by the elastic rule and indicates which tokens are visible (unmasked) or pruned (masked).

In Figure~\ref{fig:overview}, each attention mask has three visual elements:

\begin{center}
{\renewcommand{\arraystretch}{1.3}
\begin{tabular}{lll}
\toprule
\textbf{Visual Element} & \textbf{Meaning} & \textbf{Contribution to Span} \\
\midrule
Vertical purple stripe & Initial unmasked tokens (prefix) & 64 tokens (fixed) \\
Diagonal purple band   & Sliding window region             & Window length     \\
White cells            & Pruned tokens                     & 0                 \\
\bottomrule
\end{tabular}
}
\end{center}

As shown in the figure, each grid represents a candidate attention mask:
\begin{itemize}
  \item Vertically aligned masks represent masks generated from different elastic rules at the same input length.
  \item Horizontally aligned masks represent masks generated from the same elastic rule at different input lengths.
\end{itemize}

\subsubsection{Application of Attention Masks}

Attention masks for all heads are applied \textbf{in parallel}---each mask independently gates attention computation and the corresponding key-value (KV) cache entries. There is no overlap or sequential application between masks across different heads.

\subsection{Correlations Between Notions}

The following list summarizes the correlations between notions. Specifically:

\begin{enumerate}
  \item \textbf{\name configuration and elastic rule:} An \name configuration is the elastic rules of all attention heads of all layers in an LLM.
  \item \textbf{Elastic rule and attention Span:} Given an elastic rule parameterized by $(\alpha,\beta)$ and input length $N$, the attention span is computed using Equation~\ref{eq:rule_to_span}.
    \item \textbf{Attention span and attention mask:} The computed span directly determines the shape and extent of the sliding-window portion of the attention mask.
\end{enumerate}

\section{Detailed Experiment Setup}
\label{sec:appendix/experiment/experiment_setup}
\subsection{Main Setup}
\label{sec:appendix/experiment_setup}
\xhdr{Baselines}
In our experimental setup, we adhere to the specific configurations outlined in the respective papers.
In the case of StreamingLLM~\citep{xiao2023streamingLLM}, the initial four tokens remain unmasked, serving as the attention sink, except for the 70b model in Table~\ref{tab:model_comparison} and the super long setting in Figure~\ref{fig:experiment/long_long_context}, where we use 64 tokens as the attention sink.  For InfLLM~\citep{xiao2024infllm}, we adhere to the original configuration by maintaining the same local window size and selected memory size, using 128 initial tokens as specified in their setup. For H2O~\citep{Zhang2023H2O}, we ensure the same number of heavy hitter tokens and recent tokens.  Note that H2O uses original dense prefill since it relies on the column sum of the attention matrix to calculate the importance of every token for KV-Cache eviction. StreamingLLM, InfLLM and \name use efficient prefill.
Notably, in \name and StreamingLLM, the KV-Cache length equals the attention span during the efficient prefill stage. In contrast, H2O uses the original dense prefill. Besides, H2O and InfLLM require additional computations to dynamically determine the KV-Cache.

\xhdr{Models and Benchmarks}
For long-context retrieval, we use LongEval~\citep{lmsys2023longeval} to test key-value retrieval accuracy with 100 data items per length level. 
For long-context understanding, we use LV-Eval~\citep{yuan2024lveval} and LongBench~\citep{bai2023longbench}, which include 11 and 13 sub-datasets, respectively.
Since vicuna-7b-v1.5-16k and vicuna-13b-v1.5-16k~\citep{vicuna2023} can only take in 16k context length, we use the 16k split of LV-Eval benchmark~\citep{yuan2024lveval}, truncating the input to 15500 for model input in Table~\ref{tab:model_comparison}. 
For the LongBench benchmark~\citep{bai2023longbench}, we use the LongBench-E split, which features a balanced number of data items at every length level.
The LongBench dataset is segmented into ranges of 0-4k, 4-8k, and 8k+ tokens. We test each split using the input length truncation thresholds of 3,500, 7,500, and 15,500 tokens, respectively. 
Efficiency experiments measure the decode throughput on a single A100-80GB GPU at maximum batch sizes of respective methods.

\xhdr{Perplexity Evaluation}
We construct a comprehensive yet concise test set by sampling $50\times4$ data items for each length level from the test split of four long-context understanding datasets: Qasper~\citep{Dasigi2021Qasper}, MultiNews~\citep{Fabbri2019MultiNews}, TREC~\citep{li-roth-2002-trec1,hovy-etal-2001-trec2} and LCC~\citep{lcc}, representing the question answering, summarization, few-shot learning, and code completion abilities of the LLM.
Following LongBench, the data items are organized as question-answer pairs. The questions and answers are written by humans and come with the dataset.
The perplexity is calculated solely on the answer part of the data, demonstrating the model's coherence in responding to user requests.

\xhdr{Validation Dataset}
The validation dataset is used to select the \name configuration among the Pareto front solutions during the optimization step. 
The validation dataset is similarly constructed as the perplexity test dataset, but on the respective validation split of the datasets. 
$50\times4$ data items are sampled from the same four long-context understanding datasets: Qasper~\citep{Dasigi2021Qasper}, MultiNews~\citep{Fabbri2019MultiNews}, TREC~\citep{li-roth-2002-trec1,hovy-etal-2001-trec2} and LCC~\citep{lcc}. 
The additional $50$ data items from the LongEval~\citep{lmsys2023longeval} dataset are also added to validate the retrieval ability.
For the datasets that do not contain the validation split, namely TREC, MultiNews and LCC, we sample from the test split and ensure different data items with the perplexity evaluation dataset.

\xhdr{\name Settings}
\name uses the block sliding-window attention pattern with a block size of 64, where each grid depicted in Figure~\ref{fig:overview}(a) represents a block. The first block of tokens is not masked and serves as the attention sink.
For the profiling stage, we use the MultiNews~\citep{Fabbri2019MultiNews} calibration dataset with the model response as supervision, as described in Section~\ref{sec:dataset}. We use $50\times3$ data items at 2k, 4k, and 8k lengths. The data items are padded to their corresponding length level to ensure a unified shape of attention-influence tensors for each level. We adopt block granularity during profiling, calculating the average attention influence within each block to represent the block's overall influence.
For hyperparameter search space $\alpha$ and $\beta$, we use 6 values for $\alpha$ and 9 values for $\beta$, creating a search space of 54 pairs for each attention head. $\alpha$ is uniformly sampled from the range $[-2048, 8192]$, and $\beta$ is uniformly sampled from $[0,1]$. The resulting attention span lengths are clipped to the range between 0 and the current input length.
The optimization is done with the multi-objective optimization at the same set of lengths. We limit the number of distinct rules to at most two per model layer to ensure inference-time efficiency.
Among the Pareto front solutions, we select the one with the lowest perplexity on the validation dataset of length 12k.

\subsection{Efficiency Experiment Setup}
\label{sec:appendix/effiency_experiment_setup}
We test the efficiency of different frameworks using a single NVIDIA A100-SXM4-80GB GPU.
To improve the runtime profiling accuracy, we first run five forward passes as warmups. Then we use \verb|torch.CudaEvent| to calculate the runtime for each method. 
Our experiments are structured around three scenarios: including prefilling 3k tokens and decoding 1k tokens; prefilling 6k tokens and decoding 2k tokens; prefilling 12k tokens and decoding 4k tokens.
The labels are marked by the total sequence length, which equals prefill length plus decode length.

For \name, the implementation is based on HuggingFace Transformers. 
During the prefill stage, we use the sliding-window attention CUDA kernel that we designed with a block size of 64. During the decode stage, we modify the KV-Cache implementation to support our heterogeneous elastic rules. Thanks to our fixed sliding-window span during decoding, we simply replace the old KV-Cache that exceeds the span with the latest KV-Cache. Our custom decoding CUDA kernel then handles KV-Cache with varying lengths across different attention heads during decoding.

For H2O, we use its official efficient implementation, which is based on FlexGen~\citep{sheng2023flexgen}. Note that H2O uses dense prefill since it relies on the column sum of the attention matrix to calculate the importance of every token for KV-Cache eviction, which requires the attention matrix to be explicitly calculated. This makes H2O's prefill stage currently incompatible with kernel optimizations like FlashAttention. 
Therefore, H2O is prone to OOM (Out-Of-Memory) with large prefill lengths and increased batch sizes.

In our efficiency tests across all frameworks, we implemented a simple optimization at the language modeling head (lm head) during the prefill stage. Specifically, after the final layer of the transformers, we compute the logits—these are the raw outputs that are transformed into probabilities—for only the last token. This selective computation avoids generating these probabilities for preceding tokens, substantially reducing both computational overhead and memory usage. 
We also set the environment variable \texttt{PYTORCH\_CUDA\_ALLOC\_CONF} to \texttt{expandable\_segments:True} for HuggingFace and \name to mitigate memory fragmentation, allowing larger inference batch sizes.

Following the performance experiments, we use Vicuna-7B and Vicuna-13B for efficiency tests whenever possible. However, the official efficient implementation of H2O based on Flexgen only supports OPT~\citep{zhang2205opt}. Therefore, we use OPT-6.7b and OPT-13b models for H2O in Table~\ref{tab:efficiency} for comparison.

\subsection{Ablation Study Setup}
\label{sec:appendix/experiment_setup/ablation}
In the ablation study in Table~\ref{tab:dataset} and Table~\ref{tab:mask_design}, we use 25\% density instead of the 50\% used in the main experiment in Table~\ref{tab:model_comparison}. This decision is based on the observation that at a density of 50\%, the performance of the various designs is quite similar, making it difficult to discern significant differences. In contrast, a lower density of 25\% reveals more pronounced disparities between the designs, providing a clearer basis for comparison. 

In the calibration dataset experiments in Table~\ref{tab:dataset}, we intentionally exclude the influence of the validation dataset. 
We avoid using the validation dataset by profile and optimize solely at 8k length, reducing the multi-objective optimization problem to a single-objective one with only one optimal \name configuration instead of a set of Pareto fronts. 

\subsection{Input Format And Examples}
We list the prompt format and input examples used in our primary experiments and datasets. 
Dashed lines are included only for illustration clarity and are not part of the texts given to the LLMs.

\begin{bluebox}[LongEval]{label=box:longeval}

Below is a record of lines I want you to remember. Each line begins with 'line <line index>' and contains a '<REGISTER\_CONTENT>' at the end of the line as a numerical value. For each line index, memorize its corresponding <REGISTER\_CONTENT>. At the end of the record, I will ask you to retrieve the corresponding <REGISTER\_CONTENT> of a certain line index. Now the record start: 
\tcbline
line delightful-incandescence: REGISTER\_CONTENT is <19147> 

line \textbf{cloistered-presence}: REGISTER\_CONTENT is \textbf{<8862>}

...
\tcbline

Now the record is over. Tell me what is the <REGISTER\_CONTENT> in line \textbf{cloistered-presence}? I need the number.

\end{bluebox}

Format~\ref{box:longeval} illustrates the input format for the LongEval~\citep{lmsys2023longeval} retrieval benchmark. The instruction indicating which line to retrieve is provided after a lengthy context containing massive lines of register contents to remember.

\begin{bluebox}[Needle-In-A-Haystack (NIAH)]{label=box:NIAH}

People who are powerful but uncharismatic will tend to be disliked. Their power makes them a target for criticism that they don't have the charisma to disarm. That was Hillary Clinton's problem. 

\textbf{The best thing to do in San Francisco is eat a sandwich and sit in Dolores Park on a sunny day.}

It also tends to be a problem for any CEO who is more of a builder than a schmoozer.

...

\tcbline

What is the best thing to do in San Francisco?

\end{bluebox}

Format~\ref{box:NIAH} depicts the input format for another common retrieval benchmark, Needle-In-A-Haystack (NIAH)~\citep{kamradt2024NIAH}. 
The NIAH test comprises a single "needle" sentence that commonly does not fit into an irrelevant context. The model tries to answer the question based on this needle sentence.

\begin{bluebox}[MultiNews Calibration Dataset]{label=box:multinews}

You are given several news passages. Write a one-page summary of all news.

<News1>

<News2>

...

Now, write a one-page summary of all the news. 

\tcbline

\textbf{Summarization}

\end{bluebox}

Format~\ref{box:multinews} demonstrates the input format for our calibration dataset. 
The long-context MultiNews dataset~\citep{Fabbri2019MultiNews} consists of multiple news documents. 
The context includes a prompt instructing the original dense model to generate a summarization for these news articles, reflecting long-range dependencies and model alignment. 
The generated summarization serves as supervision during the cross-entropy loss calculation at the profiling stage.

\section{Additional Experiment Results}
\label{sec:appendix/additional_experiment_results}
\subsection{Performance}

\subsubsection{Overall Performance}
\label{sec:appendix/experiment/overall_performance}

\begin{table}[ht]
    \centering
    \setlength\tabcolsep{5pt}
    \begin{tabular}{ll|ccc|c|ccc|c}
    \toprule
     ~ & ~ & \multicolumn{3}{c|}{Retrieve Acc. $\uparrow$} & LV-Eval $\uparrow$ & \multicolumn{3}{c|}{LongBench $\uparrow$} & PPL $\downarrow$ \\
     Model & Attention & 4k & 8k & 16k & 16k & 0-4k & 4-8k & 8-16k & 8-12k \\
    \midrule
    \multirow{2}{*}{Vicuna-7B} & StreamingLLM & 0.91 & 0.35 & 0.09 & 4.30 & 36.39 & 32.44 & 31.04 & 3.92 \\
     & MoA & \textbf{1.00} & \textbf{0.97} & \textbf{0.58} & \textbf{5.67} & \textbf{38.07} & \textbf{33.80} & \textbf{31.75} & \textbf{3.78} \\
    \midrule
    \multirow{2}{*}{Vicuna-13B} & StreamingLLM & 0.73 & 0.81 & 0.37 & \textbf{5.65} & 36.77 & 34.65 & 33.43 & 3.70 \\
     & MoA & \textbf{0.99} & \textbf{0.97} & \textbf{0.42} & 5.57 & \textbf{41.85} & \textbf{39.76} & \textbf{36.06} & \textbf{3.62} \\
    \midrule
    \multirow{2}{*}{Llama3-8B} & StreamingLLM & \textbf{1.00} & 0.83 & 0.76 & 14.89 & 42.45 & 40.62 & 42.51 & \textbf{4.51} \\
     & MoA & 0.99 & \textbf{1.00} & \textbf{0.93} & \textbf{15.61} & \textbf{43.51} & \textbf{43.16} & \textbf{43.58} & 4.53 \\
    \bottomrule
    \end{tabular}
    \caption{Comparative analysis of retrieval accuracy, LV-Eval scores, LongBench scores, and perplexity for various models with different attention methods. All methods employ 75\% density in both prefill and decode stages.}
    \label{tab:main_table_extend}
\end{table}

Table~\ref{tab:main_table_extend} shows the overall performance of \name at a higher density of 75\%. \name shows improved performance over the baseline with the uniform attention baseline.
The progressive change of performance with respect to different densities is also shown in Figure~\ref{fig:retrieve_overall}(b) and Figure~\ref{fig:progressive_lv_eval}

\subsubsection{Long-Context Retrieval}
\label{sec:appendix/experiment/retrieval}

\begin{figure}[tb]
    \centering
    \begin{minipage}{0.45\textwidth}
        \centering
        \includegraphics[width=\textwidth]{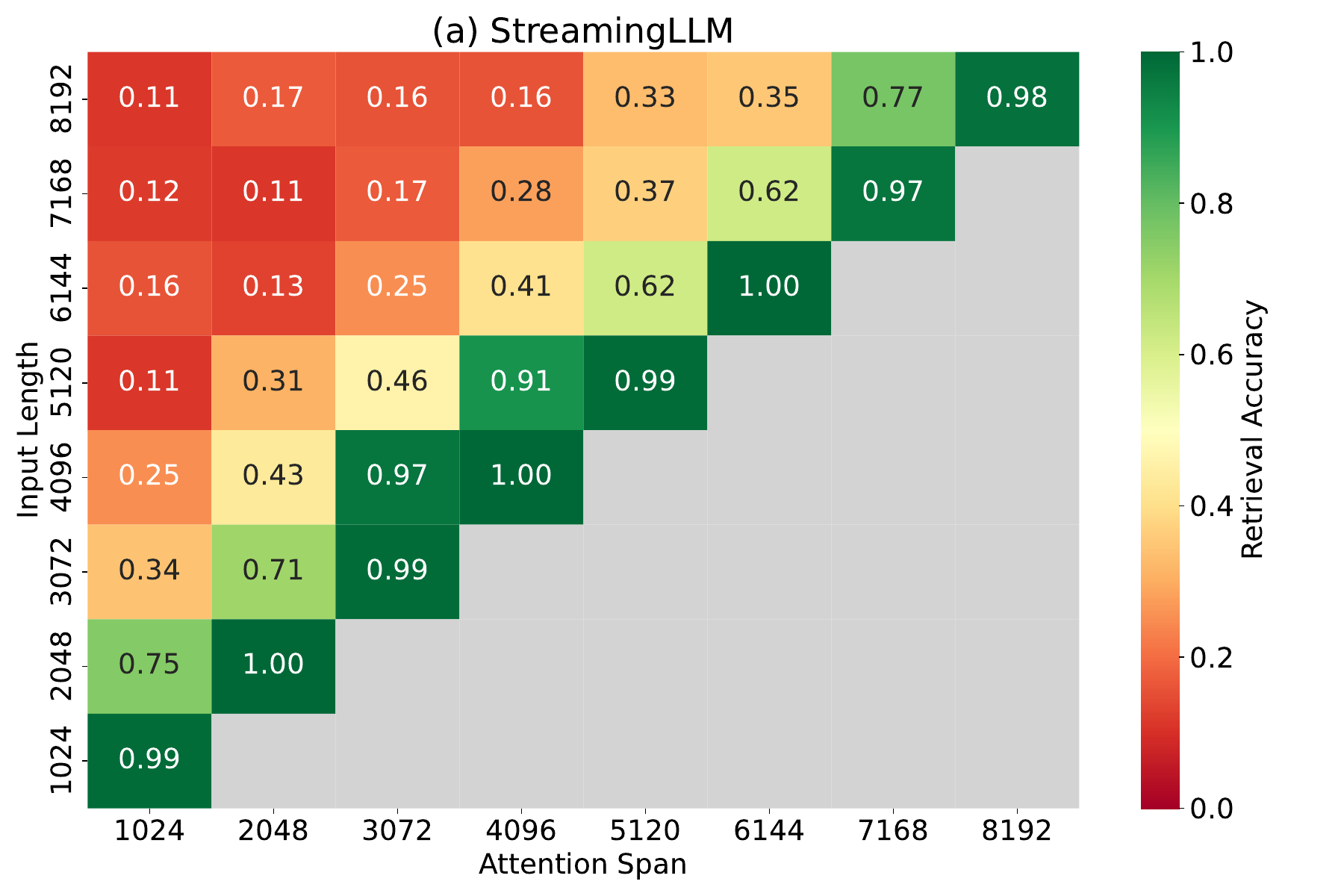} 
    \end{minipage}\hfill
    \begin{minipage}{0.45\textwidth}
        \centering
        \includegraphics[width=\textwidth]{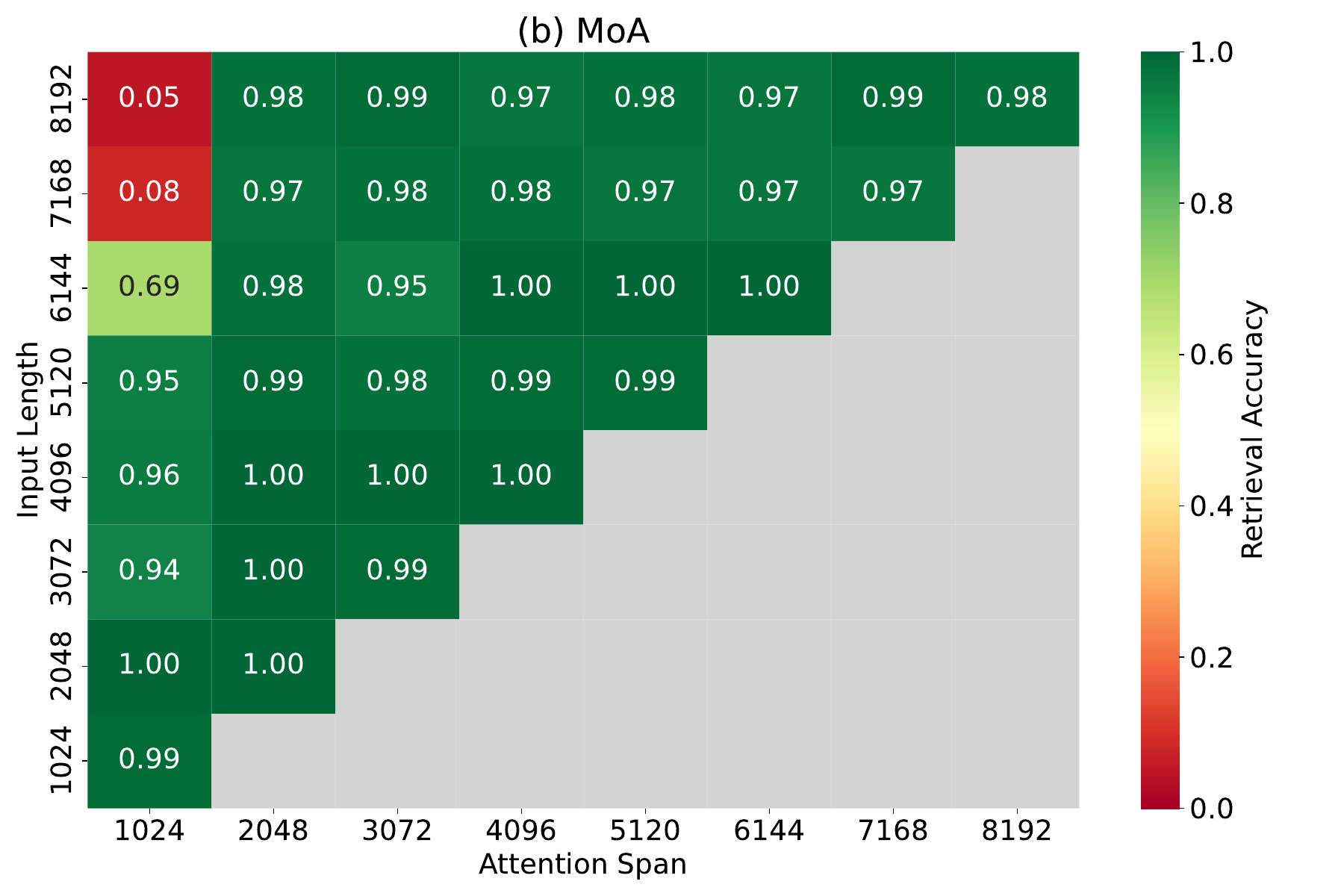} 
    \end{minipage}
    \caption{Retrieval accuracy of Vicuna-7B model using different attention methods across varying attention spans and input lengths. The X-axis shows different attention spans; the Y-axis shows different input lengths for the retrieval task. Subfigure (a) shows results for StreamingLLM, and subfigure (b) for \name.}
    \label{fig:appendix/effective_context_length_streaminglm_and_MoA}
\end{figure}

\begin{figure}[tb]
    \centering
    \includegraphics[width=0.7\textwidth]{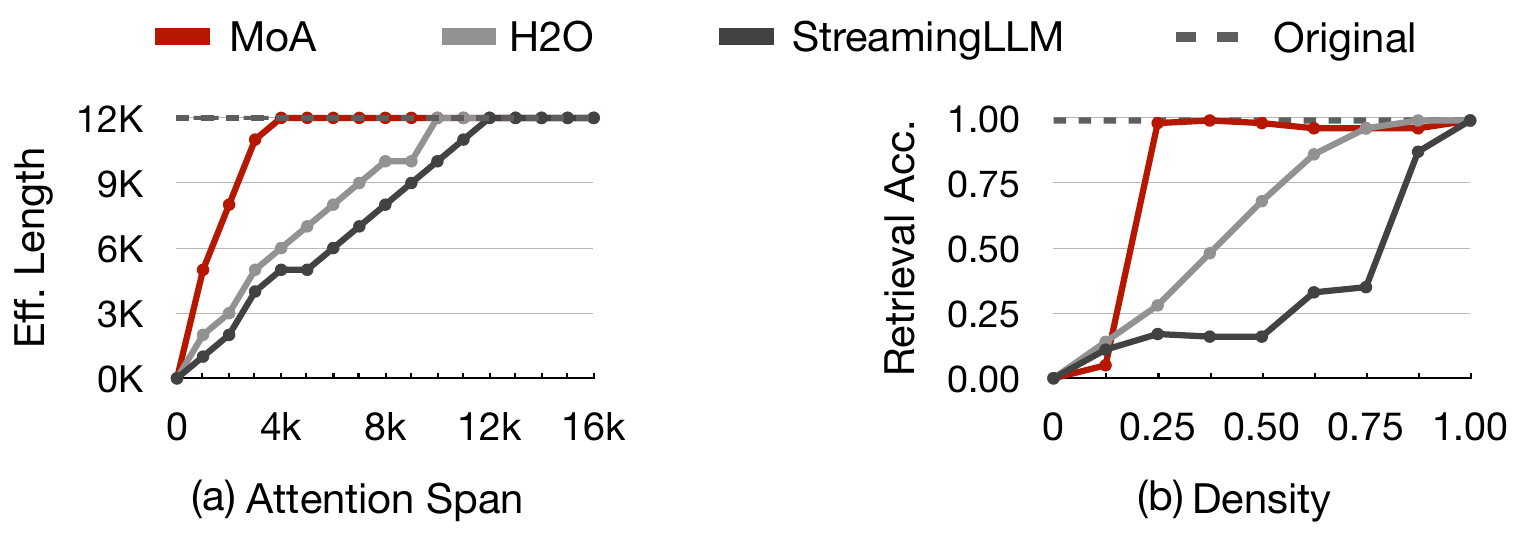}
    \vspace{-5pt}
    \caption{Retrieval accuracy tests on LongEval with Vicuna-7B. (a) Varies input lengths and densities to show effective context lengths across attention spans, 
    (b) Set input length at 8k and show retrieval accuracy across different densities.}
    \label{fig:retrieve_overall}
    \vspace{-5pt}
\end{figure}

\xhdr{LongEval Retrieval} We conduct a detailed experiment to test the retrieval ability of different attention methods across various attention spans and input lengths with the LongEval~\citep{lmsys2023longeval} dataset. 

Figure~\ref{fig:appendix/effective_context_length_streaminglm_and_MoA} shows the detailed data for effective context length calculation.
As shown in the figure, StreamingLLM can hardly maintain retrieval accuracy when the input length is beyond the attention span, while \name can effectively extend the effective context length.

Following previous work~\citep{Chen2023PI, Tworkowski2023FocusedTrans}, we quantify effective context length as the maximum input length where retrieval accuracy remains above a 90\% threshold. As shown in Figure~\ref{fig:retrieve_overall}(a), StreamingLLM and H2O achieve effective context lengths of no more than 2k tokens beyond their attention spans. In contrast, \name expands its effective context length to approximately $3.9\times$ its attention span before reaching up to the 12k limit of the original model. 
Figure~\ref{fig:retrieve_overall}(b) further shows that at a fixed input length of 8k, \name reaches over 0.9 retrieval accuracy with just 25\% density, whereas StreamingLLM and H2O require 100\% and 75\% density, respectively.

\begin{figure}[tb]
    \centering
    \includegraphics[width=0.9\textwidth]{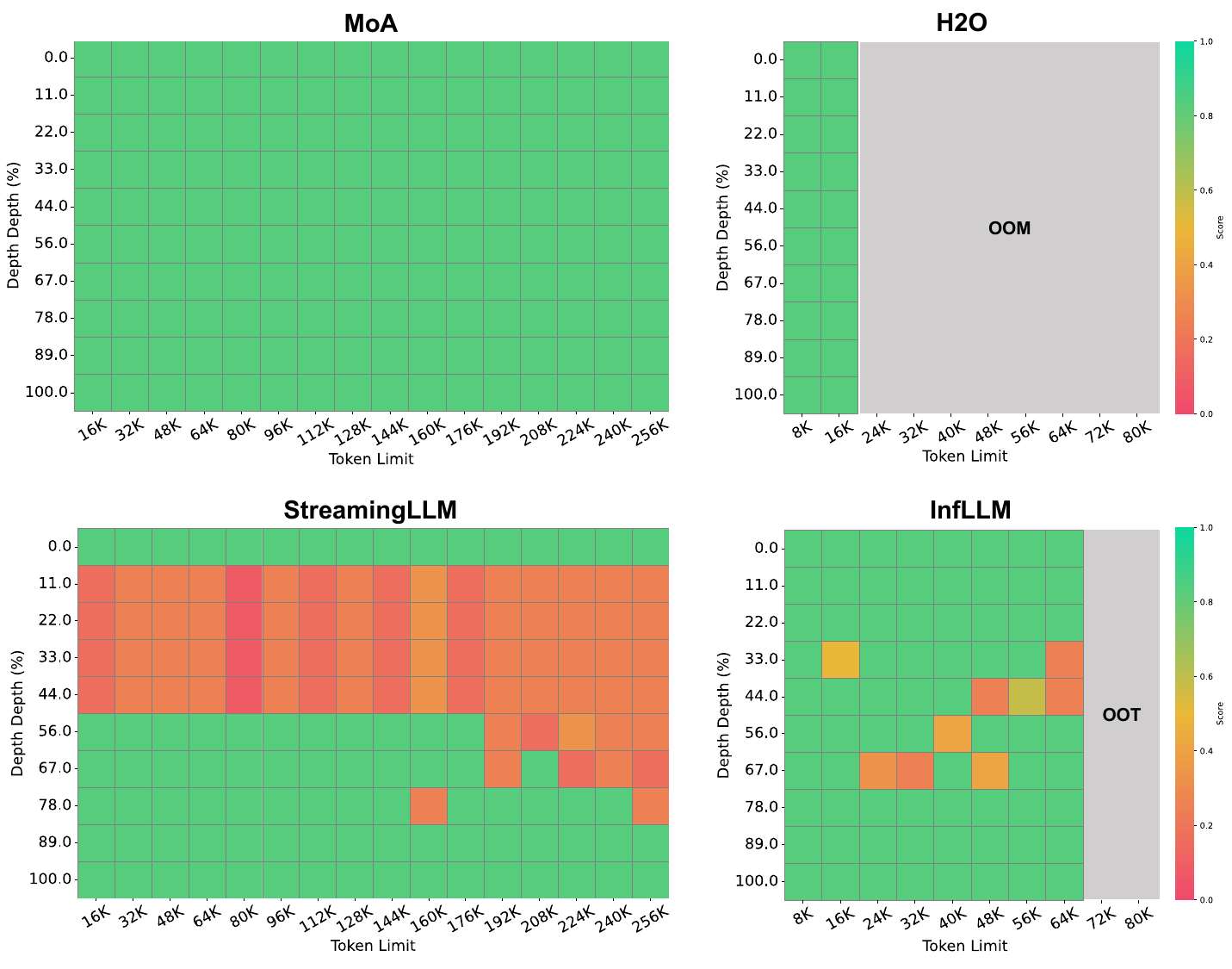}
    \caption{
    The Needle-In-A-Haystack (NIAH) retrieval accuracy using different attention methods across 8k to 256k input lengths on Llama-3-8B model. All efficient attention methods employ a 50\% density.
    }
    \label{fig:NIAH}
\end{figure}

\xhdr{Needle-In-A-Haystack (NIAH) Retrieval}
We also conduct the retrieval task using the Needle-In-A-Haystack (NIAH) dataset~\citep{kamradt2024NIAH}. 
As shown in Figure~\ref{fig:NIAH}, \name achieves perfect retrieval accuracy across input lengths ranging from 8k to 256k. 
In comparison, StreamingLLM demonstrates a limited effective context length, while InfLLM exhibits reduced retrieval accuracy within 64k input lengths.
Notably, H2O and InfLLM are unable to complete tests at extreme lengths due to Out-Of-Memory and Out-Of-Time errors.
These findings align with the results observed in the LongEval benchmark throughout the paper.

\subsubsection{Long-Context Understanding}
\label{sec:appendix/experiment/long_context_understanding}

\begin{figure}[tb]
    \centering
    \includegraphics[width=0.7\textwidth]{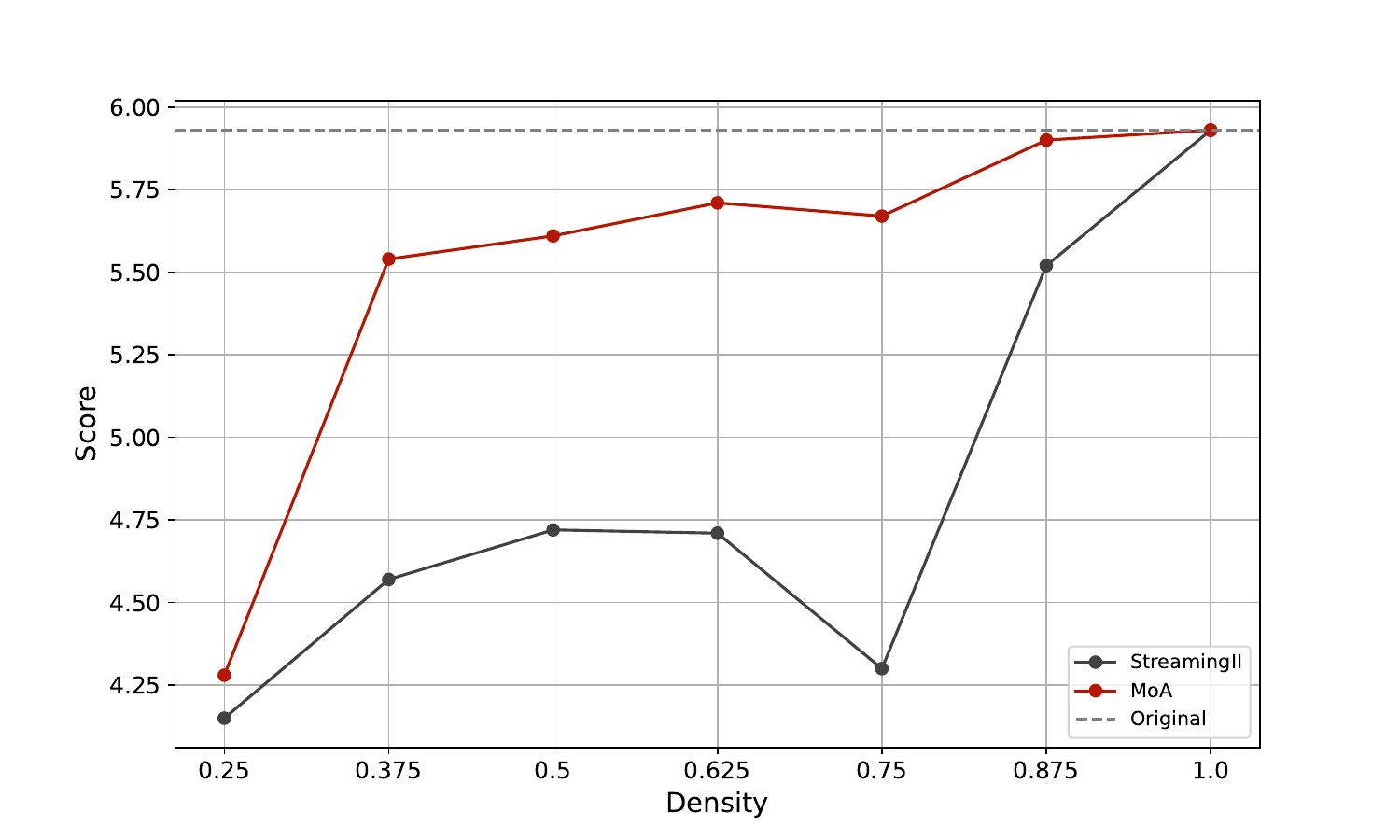}
    \caption{
    LV-Eval score of StreamingLLM and \name at various densities on Vicuna-7B model.
    }
    \label{fig:progressive_lv_eval}
\label{fig:lveval_benchmark}
\end{figure}

We conduct experiments with various densities on the LV-Eval benchmark~\citep{yuan2024lveval}. 
As shown in Figure~\ref{fig:lveval_benchmark}, \name constantly outperforms the uniform static attention baseline StreamingLLM at various densities, demonstrating the effectiveness of our heterogeneous elastic rules.

\begin{table}[tb]
    \centering
    \begin{tabular}{ll|c|c|c}
    \toprule
    ~ & ~ & \multicolumn{3}{c}{LongBench $\uparrow$}\\
    Model & Attention & 0-4k & 4-8k& 8-16k\\
    \midrule
    \multirow{5}{*}{Vicuna-7B}& Original & 37.91 & 33.82 & 32.54 \\
    \cmidrule(lr){2-5}
    & H2O & 36.23 & 32.74 & 31.81 \\
    & InfLLM & 35.23 & 33.54 & 30.15 \\
    & StreamingLLM & 30.53 & 33.28 & 31.70 \\
    & MoA & 37.04 & 32.90 & 31.94 \\
    \midrule
    \multirow{5}{*}{Vicuna-13B} & Original & 42.25 & 39.52 & 35.93 \\
    \cmidrule(lr){2-5}
    & H2O & 41.63 & 38.02 & 34.75 \\
    & InfLLM & 39.36 & 37.66 & 34.36 \\
    & StreamingLLM & 30.65 & 33.07 & 32.68 \\
    & MoA & 41.73 & 38.88 & 35.69 \\
    \midrule
    \multirow{5}{*}{Llama3-8B} & Original & 44.27 & 43.53 & 43.26 \\
    \cmidrule(lr){2-5}
    & H2O & 43.46 & 43.01 & 42.50 \\
    & InfLLM & 42.78 & 42.69 & 41.81 \\
    & StreamingLLM & 37.20 & 38.02 & 39.43 \\
    & MoA & 43.07 & 42.75 & 43.09 \\
    \midrule
    \multirow{4}{*}{Llama3-70B} & Original & 50.70 & 48.05 & 48.55 \\
    \cmidrule(lr){2-5}
    & H2O & 50.16 & 47.77 & OOM \\
    & StreamingLLM & 45.14 & 42.40 & 40.04 \\
    & MoA & 49.74 & 46.80 & 46.84 \\
    \bottomrule
    \end{tabular}
    \caption{LongBench scores for various models with different attention methods. All methods employ 50\% density in the decode stage.}
    \label{tab:longbench_score_breakdown}
\end{table}

\begin{table}[tb]
    \centering
    \footnotesize
    \setlength\tabcolsep{3pt}
    \begin{tabular}{ll|ccccc}
    \toprule
     &~ & \multicolumn{2}{c}{Single-QA} & \multicolumn{2}{c}{Multi-QA} & Retrieval\\
     Model & Attention & w/o. Conf (2) & w. Conf (2) & w/o. Conf (3) & w. Conf (2) & w. Conf (2) \\
    \midrule
     \multirow{5}{*}{Vicuna-7B}&Original& 10.49 & 6.29 & 6.83 & 5.60 & 0.00 \\
     \cmidrule(lr){2-7} 
     & H20 & 9.16 & 6.20 & 6.44 & 4.80 & 0.00 \\
     &InfLLM & 7.11 & 6.70 & 6.07 & 4.80 & 0.00 \\
     &StreamingLLM & 7.54 & 5.90 & 5.98 & 3.56 & 0.00 \\
     &MoA & 9.98 & 6.27 & 6.16 & 5.31 & 0.09 \\
    \midrule
     \multirow{5}{*}{Vicuna-13B}&Original& 10.64 & 7.28 & 5.32 & 5.07 & 1.08 \\
     \cmidrule(lr){2-7} 
     &H20 & 9.53 & 6.54 & 5.25 & 5.36 & 1.83 \\
     &InfLLM & 10.21 & 9.35 & 6.03 & 3.19 & 2.08 \\
     &StreamingLLM & 9.05 & 5.86 & 5.37 & 3.19 & 3.70 \\
     &MoA & 11.04 & 6.93 & 5.79 & 5.84 & 6.88 \\
    \midrule
     \multirow{5}{*}{Llama3-8B}&Original& 34.05 & 19.51 & 11.41 & 17.70 & 7.84 \\
     \cmidrule(lr){2-7} 
     &H20 & 28.52 & 17.05 & 11.11 & 15.98 & 9.95 \\
     &InfLLM & 24.94 & 17.75 & 10.61 & 14.80 & 6.04 \\
     &StreamingLLM & 20.21 & 9.57 & 8.14 & 9.36 & 10.03 \\
     &MoA & 32.98 & 20.53 & 10.65 & 17.57 & 8.98 \\
    \midrule
     \multirow{3}{*}{Llama3-70B}&Original& 44.44 & 25.02 & 16.71 & 22.86 & 17.43 \\
    \cmidrule(lr){2-7} 
    &StreamingLLM & 26.63 & 14.22 & 14.04 & 14.70 & 19.38 \\
     &MoA & 42.44 & 23.58 & 15.75 & 21.27 & 19.19 \\
    \bottomrule
    \end{tabular}
    \caption{Performance comparison across different models and attention methods with the LV-Eval dataset. The numbers in brackets indicate the number of sub-datasets for the category.}
    \label{tab:LVEval_breakdown}
\end{table}

\begin{figure}[tb]
    \centering
    \includegraphics[width=\textwidth]{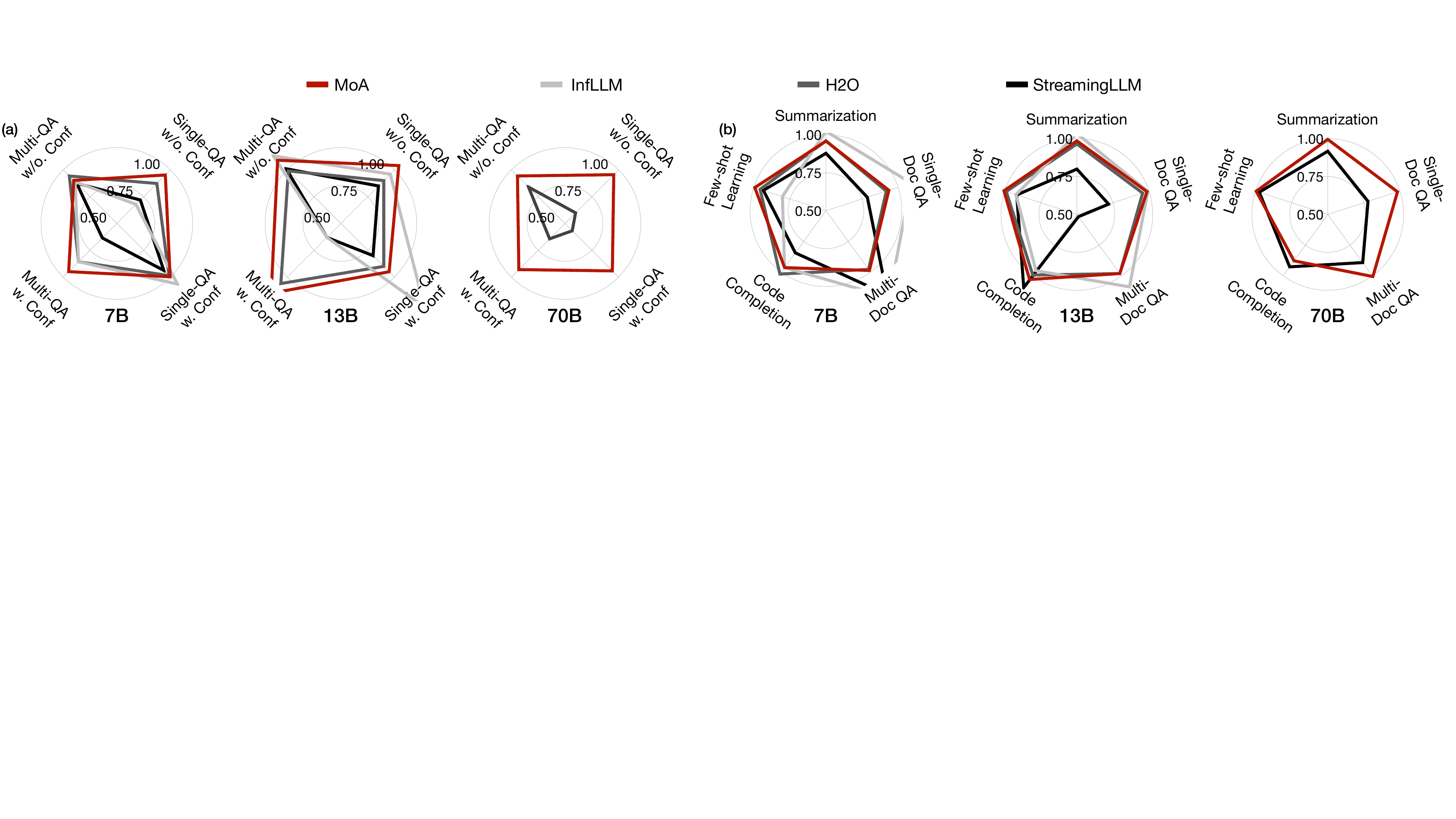}
    \caption{(a) LV-Eval and (b) LongBench scores for different attention methods at 50\% density, tested on Vicuna-7B, 13B and Llama3-70B models. Scores normalized against the original dense model.}
    \label{fig:lveval_longbench}
\end{figure}

We detailed the respective scores for LongBench and LV-Eval in Table~\ref{tab:longbench_score_breakdown} and Table~\ref{tab:LVEval_breakdown}. The number in the bracket of Table~\ref{tab:LVEval_breakdown} indicates the number of sub-datasets for the category.
\name achieves comprehensive performance comparable to the original dense model, as well as H2O that requires higher efficiency cost. 
In contrast, StreamingLLM and InfLLM display inconsistent performance: it sometimes surpasses the original model in some tasks, while suffering noticeable degradation in others.


\subsubsection{Longer-Context Generalization}

\begin{table}[ht]
    \centering
    \begin{tabular}{l|cccc}
    \toprule
    ~ & \multicolumn{4}{c}{Retrieve Acc. $\uparrow$} \\ 
    Attention  & 32k  & 64k  & 128k  & 256k \\ 
    \midrule
    SnapKV    & 1.00 & 0.88 & 0.71 & 0.33 \\ 
    PyramidKV & 1.00 & 0.85 & 0.62 & 0.37 \\ 
    MoA       & \textbf{1.00} & \textbf{0.92} & \textbf{0.83} & \textbf{0.46} \\ 
    \bottomrule
    \end{tabular}
    \caption{Retrieval accuracy at longer lengths for more recent baselines, tested at 50\% density.}
    \label{tab:scaling_context_recent}
\end{table}

We compare the retrieval accuracy with more recent works SnapKV~\citep{li2024snapkv} and PyramidKV~\citep{cai2024pyramidkv} on context lengths of 32K to 256K.
As shown in Table~\ref{tab:scaling_context_recent}, \name constantly outperforms the two latest baselines at longer contexts.

\subsubsection{Instruction-following Generation}


\begin{table}[ht]
    \centering
    \begin{tabular}{l|c|c}
    \toprule
    Attention      & Length-controlled Win Rate $\uparrow$ & Standard Error \\ 
    \midrule
    Original       & 8.84   & 0.53 \\ 
    \midrule
    H2O            & 9.66   & 0.55 \\
    InfLLM         & 5.76   & 0.42 \\ 
    StreamingLLM   & 7.96   & 0.49 \\ 
    MoA            & \textbf{9.83}   & 0.57 \\ 
    \bottomrule
    \end{tabular}
    \caption{Length-controlled win rate and its standard error of Vicuna-7B with different attention mechanisms on AlpacaEval 2.0 benchmark. All efficient attention methods employ 50\% density during decoding.}
\label{tab:alpaca_eval}
\end{table}

We evaluate \name’s performance on general instruction-following tasks using the AlpacaEval 2.0 benchmark~\citep{alpacaEvalRepo, dubois2024alphacaEvalLength}. Following the official setup, we compare the model's output with \textit{gpt4\_turbo} using the standard \textit{weighted\_alpaca\_eval\_gpt4\_turbo} evaluator, which leverages the \textit{gpt-4-1106-preview} model.
The benchmark consists of inputs and outputs with average lengths of approximately 50 and 450 tokens, respectively. To accommodate the short input lengths while maintaining a density of around 50\% during generation, we set the expected total token length to 512 and adjust hyperparameters across all methods accordingly.

Thanks to its elastic design, \name employs the same configuration used in experiments with input lengths ranging from 4k to 256k. As shown in Table~\ref{tab:alpaca_eval}, \name achieves the highest length-controlled win rate, outperforming both efficient attention baselines and the original model.


\subsection{Efficiency}
\label{sec:appendix/efficiency}

\subsubsection{Memory and Throughput Breakdown}
\label{sec:appendix/memory_throughput}

\begin{table}[th]
    \centering
    \begin{tabular}{ll|ccc}
    \toprule
    ~ & ~ & \multicolumn{3}{c}{Memory (GB)}  \\
    Size & Framework & 4k & 8k & 16k \\
    \midrule
    \multirow{3}{*}{7B} & FlashAttn2 & 28.5 & 44.4 & 76.3 \\
                        & H2O & 36.9 & OOM & OOM \\
                        & MoA & \textbf{22.7}& \textbf{32.9}& \textbf{53.5}\\
    \midrule
    \multirow{3}{*}{13B} & FlashAttn2 & 36.8 & 49.2 & 74.0 \\
                         & H2O & 40.4 & 77.9 & OOM \\
                         & MoA & \textbf{32.0}& \textbf{39.6}& \textbf{55.0}\\
    \bottomrule
    \end{tabular}
    \caption{Efficiency analysis of different frameworks on 7B and 13B models. H2O and MoA use 50\% density. GPU memory evaluated with batch sizes 8 (7B model) and 4 (13B model).}
    \label{tab:efficiency}
\end{table}

Table~\ref{tab:efficiency} highlights the memory efficiency of \name compared to H2O and FlashAttention2 on 7B and 13B models. Notably, H2O runs into Out-Of-Memory (OOM) issues at longer input lengths. In contrast, \name achieves a significant reduction in memory consumption, using $1.2$ to $1.4 \times$ less memory compared to FlashAttenion2.

We further explain the decode throughput breakdown in Table~\ref{tab:experiment/throughput}, compared to the baseline comprising Huggingface with FlashAttention2. 
The observed increase in throughput primarily stems from four aspects:

\xhdr{Static KV-Cache} \name only maintains the tokens within the span of each head,  thereby preventing growth in the KV-Cache size. This strategy eliminates the need for additional memory allocation.

\xhdr{Reduced Attention Computation} \name with features reduced density in attention span and KV-Cache. It decreases the computation and memory access required for attention computation.

\xhdr{Increased Batch Size} With the reduced size of KV-Cache, \name supports a larger batch size, contributing to the increase in throughput.

\xhdr{GPU Kernel Design} We customize \name GPU kernel using CUDA to support heterogeneous attention patterns with high efficiency.


\subsubsection{Efficiency Results for Longer Input}
\label{sec:appendix/128k_efficiency}

\begin{table}[ht]
\centering
\footnotesize
\setlength\tabcolsep{3pt} 
\begin{tabular}{lll|cc|cc}
\toprule
& & & \textbf{Min.} & \textbf{Total} & \textbf{Total} & \textbf{Throughput} \\
               \textbf{Model Size}&                    \textbf{Framework}&                    \textbf{Attention}& \textbf{\#GPU} & \textbf{Throughput} & \textbf{Memory (GB)} & \textbf{per GPU} \\
\midrule
\multirow{6}{*}{7B} & vLLM        & PagedAttention  & 2   & 30.2 & 142.0 & 15.1 \\
                    & FlexGen     & H2O             & $>$8  & -    & OOM   & -    \\
                    & HuggingFace & InfLLM          & 1   & 6.1  & 47.7  & 6.1  \\
                    & HuggingFace & StreamingLLM    & 1   & 19.8 & 43.9  & 19.8 \\
                    & HuggingFace & FlashAttention2 & 2   & 4.3  & 85.6  & 2.2  \\
                    & HuggingFace & MoA             & 1   & 20.3 & 44.0  & 20.3 \\
\midrule
\multirow{6}{*}{13B} & vLLM        & PagedAttention  & 2   & 21.5 & 142.0 & 10.8 \\
                     & FlexGen     & H2O             & $>$8  & -    & OOM   & -    \\
                     & HuggingFace & InfLLM          & 1   & 4.3  & 78.6  & 4.3  \\
                     & HuggingFace & StreamingLLM    & 1   & 14.0 & 64.6  & 14.0 \\
                     & HuggingFace & FlashAttention2 & 2   & 3.0  & 130.6 & 1.5  \\
                     & HuggingFace & MoA             & 1   & 14.7 & 63.4  & 14.7 \\
\bottomrule
\end{tabular}
\caption{Runtime efficiency at 128k input length across different methods on Vicuna-7B and 13B models. All efficient attention methods use 50\% density. Decode throughput (tokens per second) is measured with a batch size of 1, using the minimum number of A100-80GB GPUs required for testing. H2O encounters OOM error with 8 GPUs.}
\label{tab:128k_efficiency}
\end{table}

We evaluate the runtime efficiency of Vicuna-7B and 13B models at a 128k input length with a single batch size. 
Thanks to the reduced KV-Cache, MoA efficiently processes 128k input using only one A100 GPU, whereas FlashAttention2 and vLLM baselines require at least two GPUs to handle a single request. 
As shown in Table~\ref{tab:128k_efficiency}, MoA achieves a $4.7$-$4.9\times$ decode speedup compared to FlashAttention2, while using half the number of GPUs. Additionally, it demonstrates a $1.9$-$2.1\times$ reduction in GPU memory usage. 
Compared to vLLM, which utilizes tensor parallelism, MoA delivers $1.3$-$1.4\times$ higher throughput per GPU, alongside significant memory savings.

\subsubsection{Energy Efficiency Result}
We use the \texttt{pynvml} package to measure the energy consumption of GPUs. Combined with the running time, we measure the energy per token at different length levels, as shown in Table~\ref{tab:energy_efficiency}. MoA achieves an 8.7–10$\times$ reduction in energy per output token, driven by slightly lower GPU power.

\begin{table}[ht]
    \centering
    \begin{tabular}{ll|ccc|ccc}
    \toprule
    ~ & ~ & \multicolumn{3}{c}{Energy Per Token (J)} &  \multicolumn{3}{c}{Power (W)} \\
    \textbf{Framework}   & \textbf{Attention}  & 4k&8k&16k&4k&8k&16k\\ 
    \midrule
    Huggingface     & FlashAttention2   & 2.98 &5.93& 12.1   &350&354& 359 \\
    Huggingface     & MoA    & 0.34&0.62&1.21   &330&322& 315 \\
    \bottomrule
    \end{tabular}
    \caption{Per token energy consumption and average GPU power with different attention mechanisms at various sequence lengths, using Vicuna-7B on a single A100 GPU.}
\label{tab:energy_efficiency}
\end{table}

\subsubsection{Automatic Search Pipeline Overhead}
\label{sec:appendix/pipeline_cost}


\begin{table}[ht]
    \centering
    \begin{tabular}{l|ccc}
    \toprule
    Stage & 7B LLM & 13B LLM & 70B LLM \\
    \midrule
    Calibration Data Gen. & 10min & 15min & 2 $\times$ 60min \\
    Profile & 20min & 2 $\times$ 25min & 8 $\times$ 210min \\
    Optimize (CPU) & 30min & 25min & 100min \\
    Validate & 35min & 40min & 2 $\times$ 140min\\
    \midrule
    Total Latency & 1h 35min & 1h 45min & 8h 30min\\
    Total GPU Time & 1h 5min & 1h 45min & 34h 40min\\
    \bottomrule
    \end{tabular}
    \caption{Search overhead for various stages of \name across models with differing parameter sizes, reported as the amount of GPU $\times$ latency, except when only one GPU is used. Larger models necessitate more GPUs due to model parallelism. All stages utilize GPUs, except for the Optimize stage, which uses the CPU.}
    \label{tab:appendix/comression_overhead}
\end{table}

\begin{table}[ht]
    \centering
    \footnotesize
    \setlength\tabcolsep{3pt} 
    \begin{tabular}{l|c|c}
    \toprule
    Stage & \textbf{Complexity w.r.t parameter size} & \textbf{Complexity w.r.t dataset size} \\
    \midrule
    Calibration Dataset Gen. & Linear & Linear \\
    Profile & Linear & Linear \\
    Optimize & Polynomial $\sim$ Exponential for \#Head & Irrelevant \\
    Validate & Linear & Linear \\
    \midrule
    Empirical Latency & Almost Linear & Linear \\
    \bottomrule
    \end{tabular}
    \caption{Overheads for various stages of \name with respect to different parameter sizes and calibration (validation) dataset sizes.}
    \label{tab:appendix/progressive_comression_overhead}
\end{table}

We present a detailed breakdown of the time usage of \name pipeline. Table~\ref{tab:appendix/comression_overhead} summarizes the time required for various crucial phases within the \name framework, encompassing calibration dataset generation, profiling, optimization, and validation, on the Vicuna-13B model.

Profiling is the most resource-demanding part of our pipeline. For a 13b model with an 8k profile length, two A100 GPUs are required. In other cases, we only need one single GPU. Profiling on a 13b model with an 8k profile length and 50 data items takes 15 minutes. Profiling on 4k and 2k lengths takes less than 5 minutes each.

On the Intel(R) Xeon(R) Platinum 8358 2.60 GHz CPU, the optimization concludes within approximately 25 minutes. Typically, this phase generates around 10 MoA configurations. Validating each one of the configuration takes about 4 minutes, totaling around 40 minutes.

We also show the overhead of each stages in Table~\ref{tab:appendix/progressive_comression_overhead}.

\subsection{Ablation Study}

\subsubsection{Calibration Dataset}
\label{sec:appendix/experiemnt/ablation/calibration_dataset}

\begin{table}[ht]
    \centering
    \setlength\tabcolsep{3pt} 
    \begin{tabular}{lc|ccc|c}
    \toprule
    ~ & ~ & \multicolumn{3}{c|}{Test Score} & ~ \\
    Dataset & Long Dep. \& Align Model & Qasper & MultiNews & TREC &  Avg. Score \\
    \midrule
    Original & NA & 28.6 & 28.2 & 56.0 & 37.6 \\
    \midrule
    RedPajama & \xmark & 20.6 (-8.0) & 19.6 (-8.6) & \textbf{66.0} (+10.0) & 35.4 (-2.2) \\
    Qasper & \cmark & 25.6 (-3.0) & \textbf{27.8} (-0.4)& 55.0 (-1.0) & 36.1 (-1.5) \\
    MultiNews & \cmark & \textbf{29.0} (+0.4) & 27.5 (-0.7)& 54.0 (-2.0) & \textbf{36.8} (-0.8)\\
    TREC & \cmark & 27.3 (-1.3) & 27.3 (-0.9) & 55.0 (-1.0) & 36.5 (-1.1)\\
    \bottomrule
    \end{tabular}
    \caption{Performance comparison on various test sets, using different calibration sets. Tested on Vicuna-7B model. The result is tested with 50\% density \name on LongBench \citep{bai2023longbench} 0-4k split.}
    \label{tab:performance_comparison}
\end{table}

In this section, we validate the robustness of our calibration dataset design principles. We select three sub-tasks and respective datasets from the LongBench benchmark, including Qasper~\citep{Dasigi2021Qasper}, MultiNews~\citep{Fabbri2019MultiNews}, and TREC~\citep{li-roth-2002-trec1,hovy-etal-2001-trec2}. We use their training set to construct the calibration dataset, and use their respective test set in LongBench to calculate the score. Following Section~\ref{sec:dataset}, all calibration datasets are constructed using the original model's response to the context and questions as the supervision.

As shown in Table~\ref{tab:performance_comparison}, we find that as long as the calibration dataset conforms to the long-range dependency and model alignment highlighted in section~\ref{sec:dataset}, the specific choice of the dataset is less important.
Calibration datasets with long dependency and model alignment show somewhat similar test results on various datasets. Additionally, they all show strong generalization power to test sets other than their respective calibration dataset.

In contrast, the RedPajama dataset without long-range dependency and model alignment shows large variance on various test sets. It also differs from the performance of the original dense model, which may incur unexpected behaviors after the MoA search pipeline.
Note that though all datasets exhibit long dependency, the questions in the TREC dataset can be answered without long context. The context in the TREC dataset of LongBench is the many-shot examples, each showing a short sentence and its classification result, while the question is to classify a new short sentence. Although the context helps to determine the complete set of 50 classes, the model can also directly clarify the sentence without any context based on common knowledge. 
It may contribute to a high score on the TREC test set with the RedPajama calibration dataset.


\section{\name Configuration Analysis}
\label{sec:insight}
\input{content/insight}

\section{Automatic Pipeline Details}
\label{sec:appendix/pipeline_detail_derivation}

\subsection{Additional Oracle on Elastic Pattern Design}
\label{sec:appendix/oracle}

\begin{figure}[ht]
    \centering
        \includegraphics[width=0.75\textwidth]{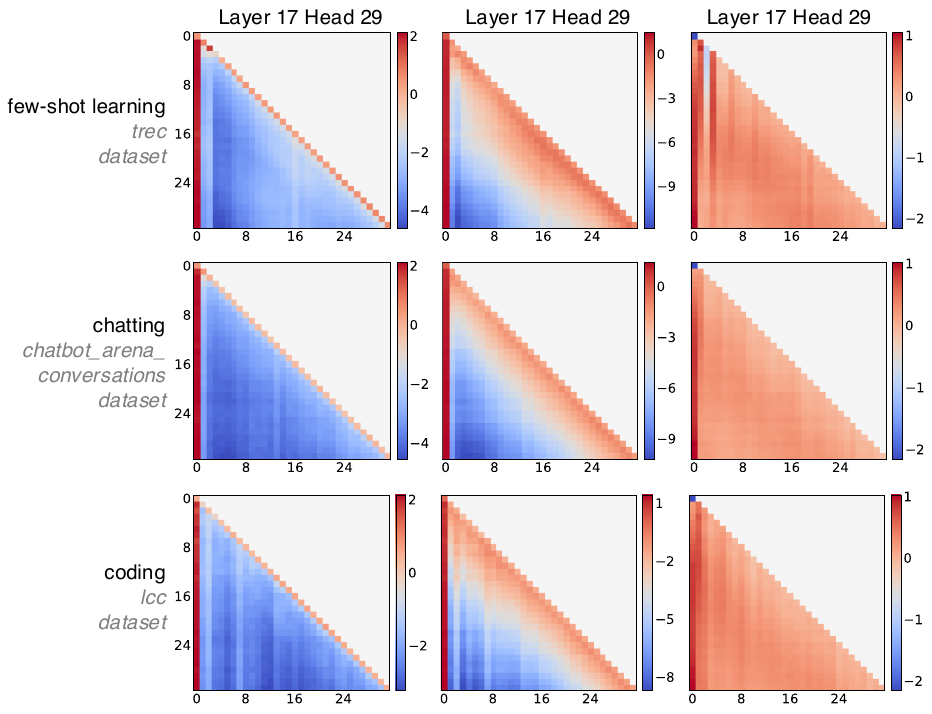}
        \caption{Examples of attention matrices from different attention heads (columns) and tasks (rows) of the Vicuna-7B model. The attention matrices were averaged over 256 data items per dataset. The same head shows a similar attention span across different tasks, explaining the robust cross-dataset generalizability of our method.}
    \label{fig:oracle_datasets}
\end{figure}

We visualize the attention matrix of the same attention heads across three additional tasks in Figure~\ref{fig:oracle_datasets}, as an extension of Figure~\ref{fig:attention_oracle}. The consistent attention span across tasks sheds light on the strong cross-dataset generalization ability of our \name method.

\subsection{Derivation of Attention Influence}
\label{sec:appendix/derive_attention_influence}

We use the first-order Taylor expansion to calculate the influence of each attention value. This approximation approach is supported by methodologies commonly employed in other LLM compression approaches~\citep{llm-mq,shi2021sparsebert,das2023beyond,jiang2023pushing}.

As discussed in Section~\ref{sec:pipeline/profile}, when masking out attention value $A_{h,i,j}$ at head $h$, row $i$, and column $j$, it also influences the attention values in the same row by $\Delta A_{h,i,n|j}$.
\begin{equation}
\begin{aligned}
    A_{h,i,n} &= \frac{e^{S_{h,i,n}}}{\sum_j e^{S_{h,i,j}}} \\
    \Delta A_{h,i,n|j} &=
    \begin{cases}
        -A_{h,i,n}, & n=j \\
        A_{h,i,n} ({\sum_j e^{S_{h,i,j}}} / {\sum_{j\neq n} e^{S_{h,i,j}}} - 1), & n \neq j\\
    \end{cases}
\end{aligned}
\label{eq:delta_A}
\end{equation}

Following the definition, the attention influence $\textbf{E}_h$ is calculated as follows:
\begin{equation}
    E_{h,i,j} = \sum_n \frac{\partial L}{\partial A_{h,i,n}} \cdot \Delta A_{h,i,n|j} 
    \label{eq:each_effect}
\end{equation}
Given Equation~\ref{eq:each_effect} and ~\ref{eq:delta_A}, we derive Equation~\ref{eq:effect} as follows. For notation simplicity, we omit the head index $h$ here.
\begin{equation}
\begin{aligned}
    E_{i,j} &= \sum_n \frac{\partial L}{\partial A_{i,n}} \cdot \Delta A_{i,n|j}  \\
    &= \frac{\partial L}{\partial A_{i,j}} \cdot (-A_{i,j}) + \sum_{n\neq j} \frac{\partial L}{\partial A_{i,n}} \cdot A_{i,n} \cdot \left(\frac{\sum_k e^{S_{i,k}}}{\sum_{k\neq j} e^{S_{i,k}}} - 1\right) \\
    &= \frac{\partial L}{\partial A_{i,j}} \cdot (-A_{i,j}) + \sum_{n\neq j} \frac{\partial L}{\partial A_{i,n}} \cdot A_{i,n} \cdot \frac{e^{S_{i,j}}}{{\sum_{k} e^{S_{i,k}}} - e^{S_{i,j}}} \\
    &= \frac{\partial L}{\partial A_{i,j}} \cdot (-A_{i,j}) + \sum_{n\neq j} \frac{\partial L}{\partial A_{i,n}} \cdot A_{i,n} \cdot \frac{e^{S_{i,j}}/{\sum_{k} e^{S_{i,k}}}}{1 - e^{S_{i,j}}/{\sum_{k} e^{S_{i,k}}}} \\
    &= \frac{\partial L}{\partial A_{i,j}} \cdot (-A_{i,j}) 
    + \sum_{n\neq j} \frac{\partial L}{\partial A_{i,n}} \cdot A_{i,n} \cdot \frac{A_{i,j}}{1 - A_{i,j}} \\
    &= \frac{\partial L}{\partial A_{i,j}} \cdot (-A_{i,j}) 
    - \frac{\partial L}{\partial A_{i,j}} \cdot A_{i,j} \cdot \frac{A_{i,j}}{1 - A_{i,j}}
    + \sum_{n} \frac{\partial L}{\partial A_{i,n}} \cdot A_{i,n} \cdot \frac{A_{i,j}}{1 - A_{i,j}} \\
    &= \frac{\partial L}{\partial A_{i,j}} \cdot \left(-\frac{A_{i,j}}{1 - A_{i,j}}\right)
    + \frac{A_{i,j}}{1 - A_{i,j}} \cdot
    \sum_{n} \frac{\partial L}{\partial A_{i,n}} \cdot A_{i,n}  \\
    &= - \frac{A_{i,j}}{1 - A_{i,j}} \left(\frac{\partial L}{\partial A_{i,j}}
    - \sum_{n} \frac{\partial L}{\partial A_{i,n}} \cdot A_{i,n} \right) \\
\end{aligned}
\end{equation}

It is worth noting to mention that it can also be formulated as matrix multiplications:
\begin{equation}
    \textbf{E}_h = 
        \frac{\textbf{A}_h}{1-\textbf{A}_h} \cdot \left( \frac{\partial L}{\partial \textbf{A}_h} - \left(\frac{\partial L}{\partial \textbf{A}_h} \cdot \textbf{A}_h\right)\mathbbm{1}^{N\times N} \right).
\end{equation}

\subsection{Optimization Details}
\label{sec:appendix/optimization_details}

\subsubsection{Optimizing at Single Length}
The optimization problem is formulated as follows:

\begin{equation}
    \mathop{\arg\min} \Delta L = \sum_h \Delta L_{h,r_h}, \quad
    \operatorname{s.t.}  \frac{1}{H}\sum_h d_{r_h} \leq d_{\text{constr}}.
    \label{eq:optimization}
\end{equation}

To transform the optimization problem into a standard Mixed-Integer Programming (MIP) framework, we introduce the binary variable $X_{h,r_h} \in \{0,1\}$. It indicates whether to select rule $r_h$ for the attention head $h$. Assume the model has $H$ attention head, and head $h$ has $R_h$ elastic rules.

\begin{subequations}
\label{eq:mip}
\begin{align}
    \mathop{\arg\min} \frac{1}{H}\sum_{h=0}^{H-1} \sum_{r_h=0}^{R_h -1} \Delta L_{h,r_h} X_{h,r_h} & \quad \operatorname{s.t.} \label{eq:mip:obj} \\
    \sum_{r_h=0}^{R_h-1} X_{h,r_h} = 1, & \quad h \in \left\{0,\cdots, H-1 \right\} \label{eq:mip:plan} \\
    \frac{1}{H}\sum_{h=0}^{H-1} \sum_{r_h=0}^{R_h -  1} d_{r_h} X_{h,r_h} \leq d_{\text{constr}} & \label{eq:mip:latency} \\
    0\leq X_{h, r_h}\leq 1, X_{h,r_h} \in \mathbb{Z}, & \quad \forall h \in \left\{0,\cdots, H-1 \right\}, \forall r_h \in \mathbb{R} \label{eq:mip:binary}
\end{align}
\end{subequations}

In this formulation, \eqref{eq:mip:obj} serves as the objective function to minimize the loss, subject to the constraints that each matrix selects exactly one \name configuration \eqref{eq:mip:plan}, and the average density does not exceed $d_{\mathrm{constr}}$ \eqref{eq:mip:latency}. Finally, \eqref{eq:mip:binary} enforces that $X_{h,r_h}$ is a binary variable, indicating the selection of plans.

Additionally, to enforce the restriction that each model layer only has a limited number of different plans, we bound the norm of element-wise multiplication of $\mathbf{X}_{h} = \begin{bmatrix}
    X_{h, 0} & X_{h, 1} & \cdots & X_{h, R_h -1}
\end{bmatrix} ^\top $ in a single layer.

\subsubsection{Optimizing at Multiple Lengths}

\begin{figure}[htb]
    \centering
        \includegraphics[width=\textwidth]{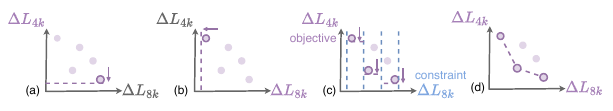}
        \caption{Illustration of our multi-objective Mixed-Integer Programming (MIP) approach, using a two-objective optimization example: 
        (a) \name first minimizes the loss for 4k inputs and records the corresponding loss for the current optimal plan at 8k.
        (b) Next, it minimizes the loss for 8k inputs and records the corresponding loss for the current optimal plan at 4k. These steps establish the loss ranges $R$ for both 4k and 8k input lengths.
        (c) \name then re-optimizes the loss at 4k, this time using the loss intervals at 8k as different constraints. All plans generated under these constraints are recorded. 
        (d) The last process (c) is repeated for 8k, using 4k intervals as constraints. Finally, plans meeting the Pareto front criteria for both 4k and 8k inputs are selected as the final outputs.}
    \label{fig:MIP}
\end{figure}

With the ability to optimize at a single length, we utilize the same framework for multi-objective MIP across various lengths. 
The key is to transform the multi-objective MIP problem into several single-objective MIP problems~\citep{Paria2018AFF}.
We utilize the idea of epsilon-constraint method~\citep{haimes1971bicriterion}.

Figure~\ref{fig:MIP} illustrates the optimization process for two input lengths. We discuss the generalized approach to handle an arbitrary number of lengths.
We first select each input length as our primary objective to perform the single-objective optimization on it while simultaneously recording the outcomes of other objectives. Specifically, for $N$ distinct objectives, we do single-objective MIP optimization on the $i$-th objective, getting minimum loss $\Delta L^{(N_i)}_i$, and we concurrently collect losses of other objectives $\Delta L^{(N_j)}_i$ for $ j \neq i$. This process allows us to establish the range of loss  $R^{(N_j)} = \left[ \min_{i}\Delta L^{(N_j)}_i , \max_{i} \Delta L^{(N_j)}_i \right]$ for each objective.
Then, we iterate through each objective again. Compared with the original multi-objective optimization in Equation~\ref{eq:multi-optimize}, we now consider other objectives as constraints. To implement this, we partition each loss range $R^{(N_j)}$ of other objectives $j \neq i$ into $M$ uniform intervals $S^{(N_j)}_{k}$, where $0 \leq k < M$. We then solve the MIP problems for each objective $i$ and iterating through the constraint intervals:
\begin{equation}
    \mathop{\arg\min}_{r_h \in \mathbb{R}} \Delta L^{(N_i)}
    \quad \operatorname{s.t.} 
    \space \frac{1}{H}\sum_{h=1}^H d^{(N_i)}_{r_h} \leq d^{(N_i)}_\text{constr}, \forall N_i \in \mathbb{N}_{\text{constr}};
    \quad
    \Delta L^{(N_j)} \in S^{(N_j)} _{k_j}, \forall j \neq i.
    \label{eq:multi_single_objective}
\end{equation}
where this optimization is performed for each $i$ ranging from $0$ to $N$.
For each $j$, $k_j$ can vary independently from $0$ to $M$.
For efficiency consideration, we set the number of intervals as five. 
Finally, the results that do not conform to the Pareto front requirements are removed, resulting in the final Pareto front set of our multi-objective optimization problem.

\subsection{Coding Interface}
\label{sec:appendix/interface}

\name offers a straightforward and intuitive interface for configuration search and inference. For supported HuggingFace LLM architectures, the entire search pipeline for any given new model is fully automated with a single command:

\begin{verbatim}
python scripts/pipeline/main.py –model_path <HF_MODEL_PATH> –model_name <NAME>
\end{verbatim}

Once search is completed, the \name configuration is as straightforward to deploy as its dense counterpart. 
\name seamlessly integrates with the HuggingFace framework and remains fully compatible with high-level methods like \texttt{pipeline} and \texttt{generate}.
Switching from standard attention to our mixture-structured sliding-window attention requires minimal adjustments. We provide a specialized CUDA GPU kernel for heterogeneous sliding-window attention, integrated into HuggingFace's \texttt{AttentionModule} interface. The adoption involves just two lines of code:

\begin{verbatim}
model = update_model_function(model, model_name)
model.model.set_mixture_of_attention(moa_config)
\end{verbatim}

By automating the \name configuration search process and offering a simple deployment interface, our approach minimizes complexity and ensures practical usability.

\section{Limitations and Future Work}
Under an extremely low-density budget, \name fails to maintain good performance. Designing a dynamic \name method has the potential to address this issue, which we leave for future work. 
Using non-linear elastic rules with bounded attention spans is also worth exploring. 
Additionally, \name's profiling method can be adapted to evaluate the influence of weights and other activations, facilitating other compression methods such as quantization.

%% file: content/insight.tex
\subsection{Overview}
\label{sec:rules-by-moa}
We investigate \name's elastic rules for each head. As shown in Figure~\ref{fig:insight/density}, masks in the initial and middle layers exhibit high density, aligning with conclusions from previous research on LLMs' intrinsic dimensions~\citep{Valeriani2023hiddenRepresent} and layer sensitivities~\citep{Yuan2023ASVD}. 
Conversely, in the final layers, most heads require low density, while few need high density. 
Figure~\ref{fig:insight/layer_density_range} shows that layers with lower average density typically display more diverse densities among heads, confirming the need for heterogeneity within the same layer. 

\subsection{Statistics on Rules Discovered by \name}

\begin{figure}[ht]
    \centering
    \begin{minipage}{0.48\textwidth}
        \centering
        \includegraphics[width=\linewidth]{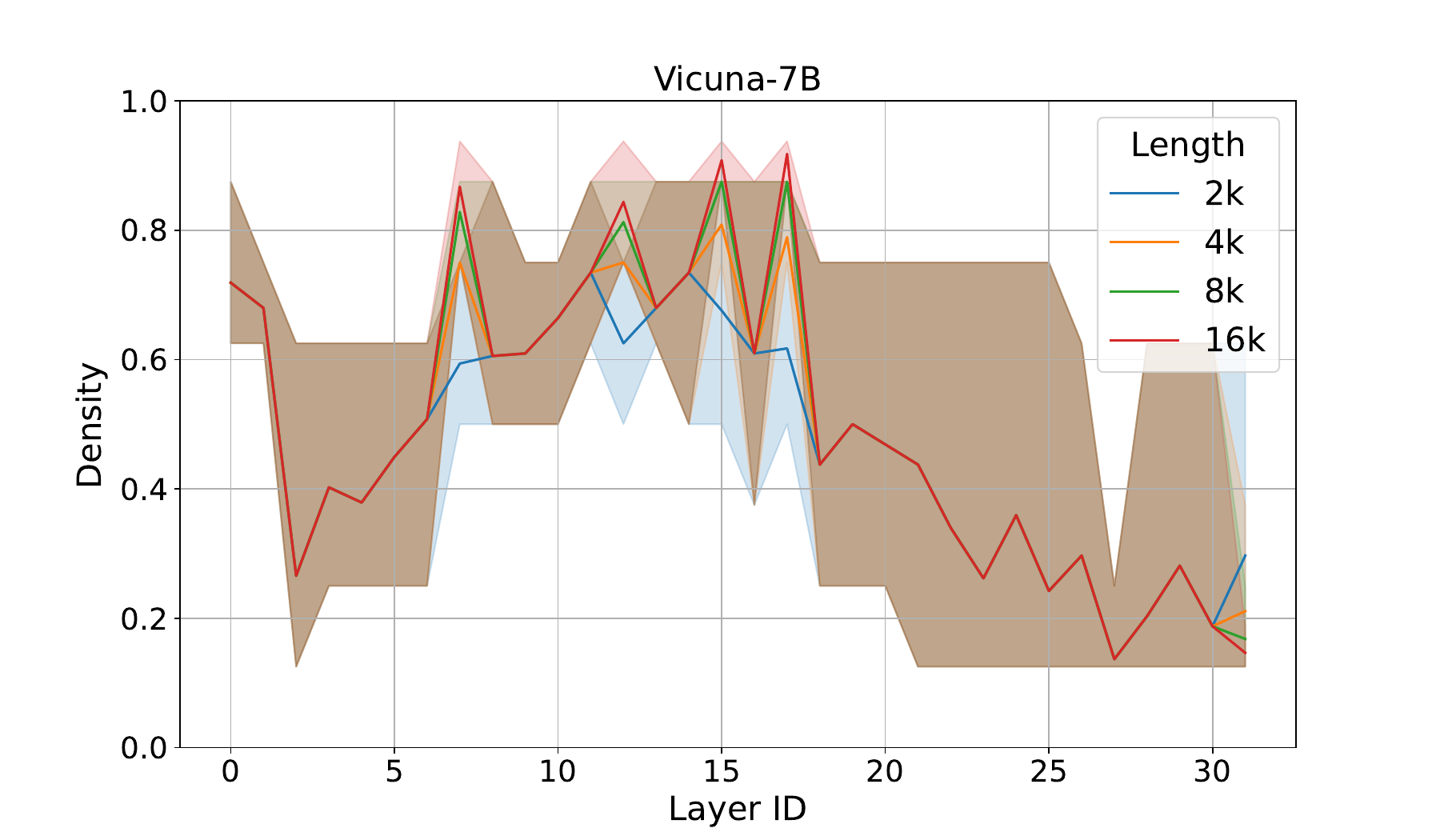}
    \end{minipage}\hfill
    \begin{minipage}{0.48\textwidth}
        \centering
        \includegraphics[width=\linewidth]{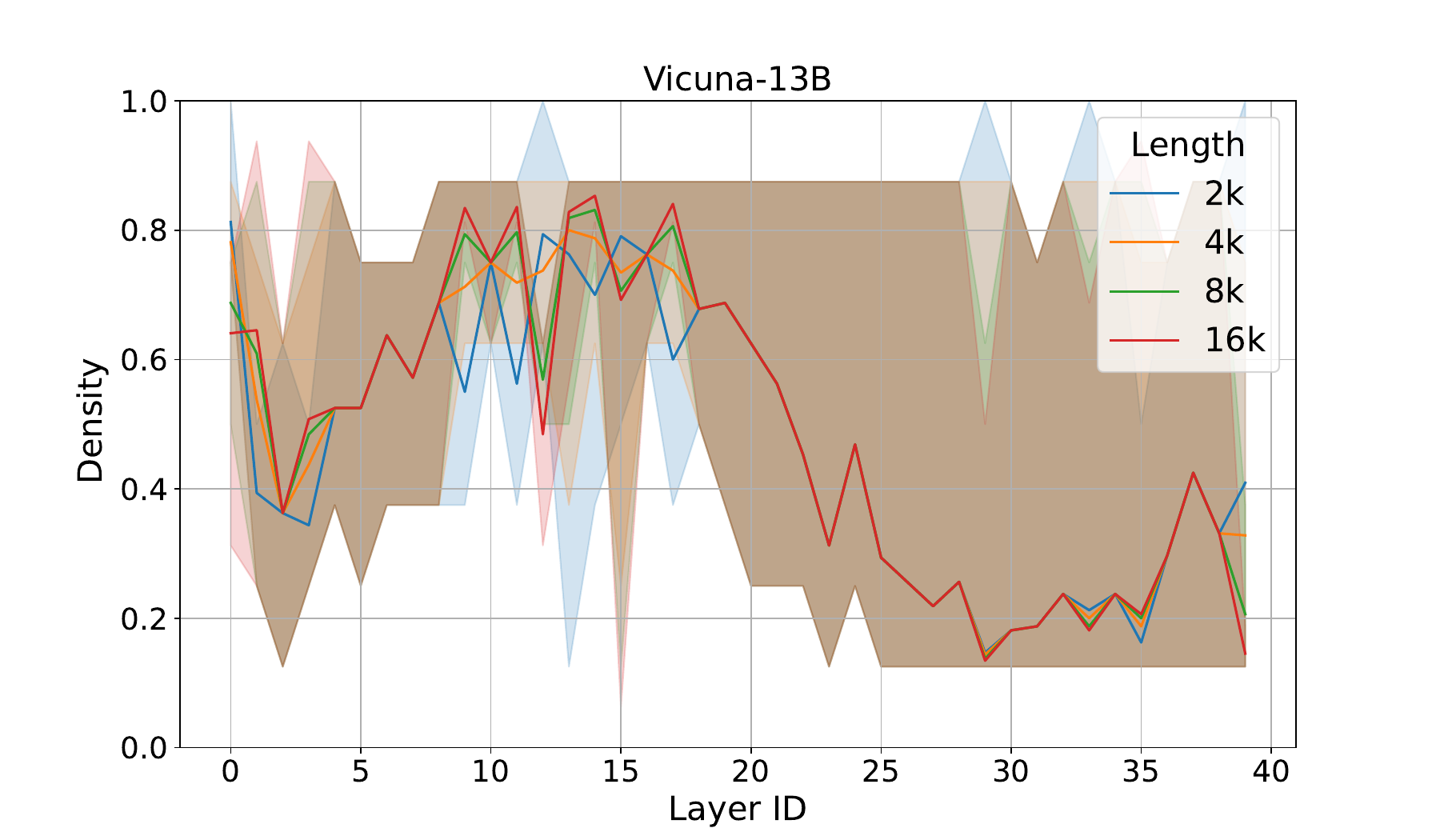}
    \end{minipage}
    \caption{The \name mask density across layers for different LLMs.}
    \label{fig:insight/density}
\end{figure}

\begin{figure}[ht]
    \centering
    \begin{minipage}{0.48\textwidth}
        \centering
        \includegraphics[width=\linewidth]{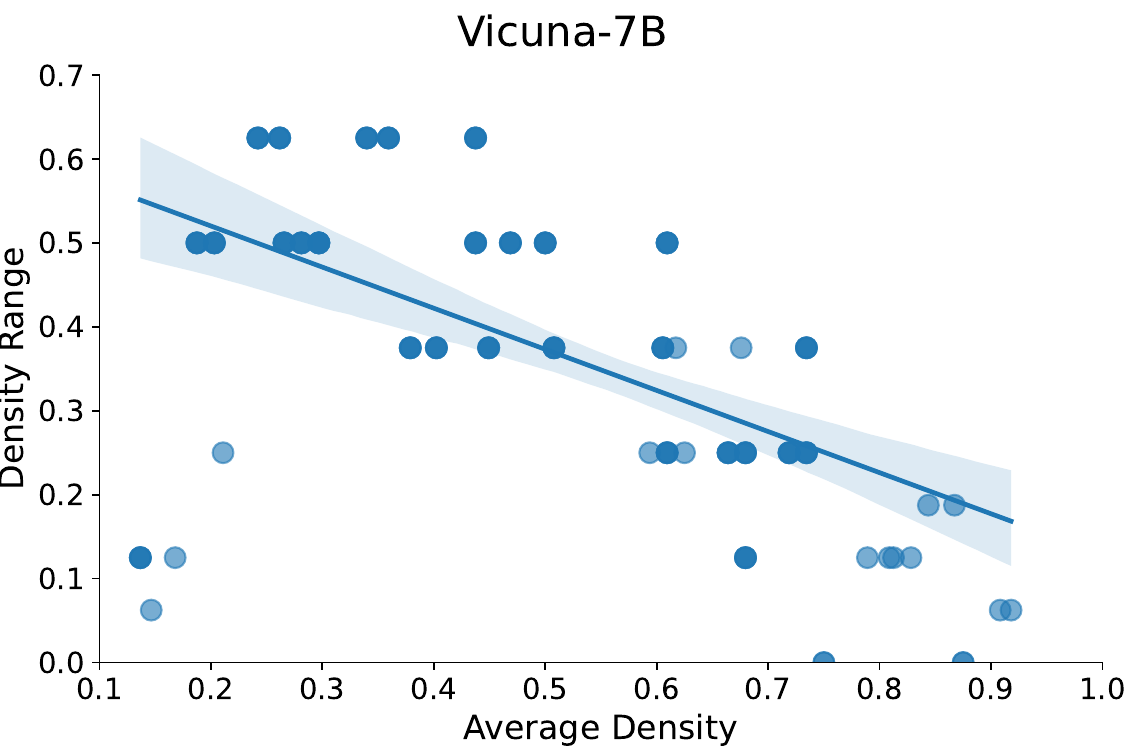}
    \end{minipage}\hfill
    \begin{minipage}{0.48\textwidth}
        \centering
        \includegraphics[width=\linewidth]{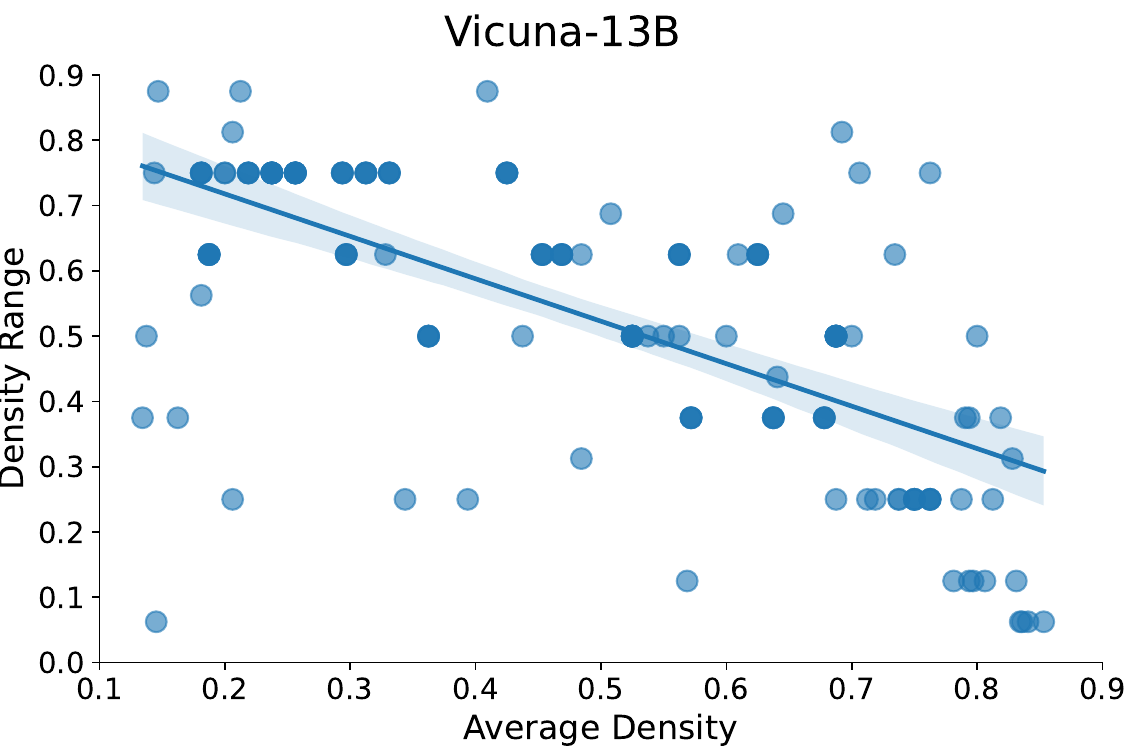}
    \end{minipage}
    \caption{The \name mask's average density and the density range for each layer for different LLMs.}
    \label{fig:insight/layer_density_range}
\end{figure}

This subsection provides empirical evidence for rules discovered by \name, as mentioned in Section \ref{sec:rules-by-moa}. The lines and spans in Figure \ref{fig:insight/density} show that all heads in the first few layers generally need a high KV-Cache density and are thus assigned larger attention spans. Subsequently, a few layers generally require only medium density. In the final layers, most heads require low density, while some outlier heads need high density. This observation conforms to previous findings about the intrinsic dimension of LLMs~\cite{Valeriani2023hiddenRepresent}. The geometry of density is similar to the intrinsic dimension, with two local minima. 
As observed in Figure \ref{fig:insight/density}, layers with lower average density (smaller values on the lines) typically display a wider range of density (wider shades). Figure \ref{fig:insight/layer_density_range} validates this observation and confirms the need for heterogeneous attention rules within the same layer. 

\subsection{Connections Between \name Rule and Semantic}

In this section, we investigate the masks acquired with \name and show their interpretable semantics. 
Previous works manually restrict the model's attention pattern, which may harm the semantics learned by the dense model.
In contrast, \name preserves semantics with statistical analysis and optimization. We use visualization, human interpretation, and quantitative methods to analyze the semantics of the original model and verify whether \name captures them.

\subsubsection{Mask Visualization and Semantic Categorization}
\label{sec:mask_vis}

Given any token, two kinds of information are used as the model inputs: positional encoding and token embedding. 
Position encoding indicates the absolute~\citep{zhang2205opt} or relative positions~\citep{touvron2023llama2} of tokens in the sentence. 
Token embedding maps different tokens as different vectors.
Attention head $h$ responds to both types of information and outputs the corresponding attention value $A_h$. 
As shown in equation~\ref{eq:token-position-based}, we denote the influence of position and token of head $h$ as function $P_h$ and $T_h$, respectively.
The attention value $A_{h,i,j}$ between the $i$th and $j$th token $t_i$ and $t_j$ is determined by the combination $f_h$ of position and token influence functions. 

\begin{equation}
    A_{h,i,j} = \mathbb{A}_h(t_i, t_j, i, j) 
    = f_h \left( P_h(i,j), T_h(t_i, t_j) \right) 
    \label{eq:token-position-based}
\end{equation}

Figure~\ref{fig:attention_oracle} visualizes two typical heads that are either dominated by position $P$ or token $T$ function.
For the first attention head in Figure~\ref{fig:attention_oracle}, local positional attention is clearly observed. In this head, regardless of the sentence, each token pays major attention to the first token and the prior token. As a result, the mean attention matrix accumulates extremely large attention values in the first column and on the sub-diagonal.
In contrast, the second attention head in Figure~\ref{fig:attention_oracle} places more emphasis on content-based attention. Since the positional distribution of important tokens is generally random, the attention matrix can show large attention values at any position, resulting in a mean attention matrix without extreme values.

In conclusion, the mean attention matrix across different sentences provides valuable insight into whether an attention head is more position-based or content-based. Intuitively, the more uneven the distribution of attention matrix values is, the more position-based the head is.

\subsubsection{Quantitative Semantic Analysis}

We quantify how position-based an attention head is and analyze whether \name successfully utilizes such semantics through the evaluate–generate–optimize pipeline. We model Equation~\ref{eq:token-position-based} with a linear approximation.
$P_h$ and $T_h$ are random variables with the same expectation $\mu$ and standard deviation $\delta$ for all heads. 
For attention head $h$, the weight factor $\alpha_h$ evaluates the relative influence of position and token on the final attention value. 

\begin{equation}
    A_{h,i,j} = \alpha_h P_h(i,j) + (1-\alpha_h) T_h(t_i, t_j)
    \label{eq:linear_attention}
\end{equation}

Given the randomness of token positions in long context, we assume that the token position and its content are irrelevant. For different sentences $s$, the expectation $\mathbb{E}_t$ of the attention value between position $i$ and $j$ can be expressed as follows. Note that it excludes the matrix diagonal since $T_h(t_i, t_j), i\neq j$ and $T_h(t_i, t_i)$ may follow different distributions.

\begin{equation}
\begin{aligned}
    \mathbb{E}_t[A_{h,i,j}] 
    &= \frac{1}{S} \sum_{s=1}^S \left(\alpha_h P_h(i,j) + (1-\alpha_h) T_h(t_i^{(s)}, t_j^{(s)}) \right) \\
    &= \alpha_h P_h(i,j) + (1-\alpha_h) \frac{1}{S} \sum_{s=1}^S T_h(t_i^{(s)}, t_j^{(s)}) \\
    &= \alpha_h P_h(i,j) + (1-\alpha_h)\mu_T , \forall i > j\\
    \label{eq:mean_map}
\end{aligned}
\end{equation}

The standard division $\sigma_p$ of $\mathbb{E}_t$ over different positions of the attention matrix is

\begin{equation}
\begin{aligned}
    \sigma_p(\mathbb{E}_t[A_{h,i,j})
    &= \sqrt{\frac{2}{(1+N)N} \sum_{i,j\in[1,N),i > j} [(\alpha_h P_h(i,j) + (1-\alpha_h)\mu_T) - (\alpha_h \mu_P + (1-\alpha_h)\mu_T)]^2 }\\
    &= \alpha_h \delta_p
    \label{eq:std_mean_map}
\end{aligned}
\end{equation}

\begin{figure}
    \centering
    \includegraphics[width=0.5\textwidth]{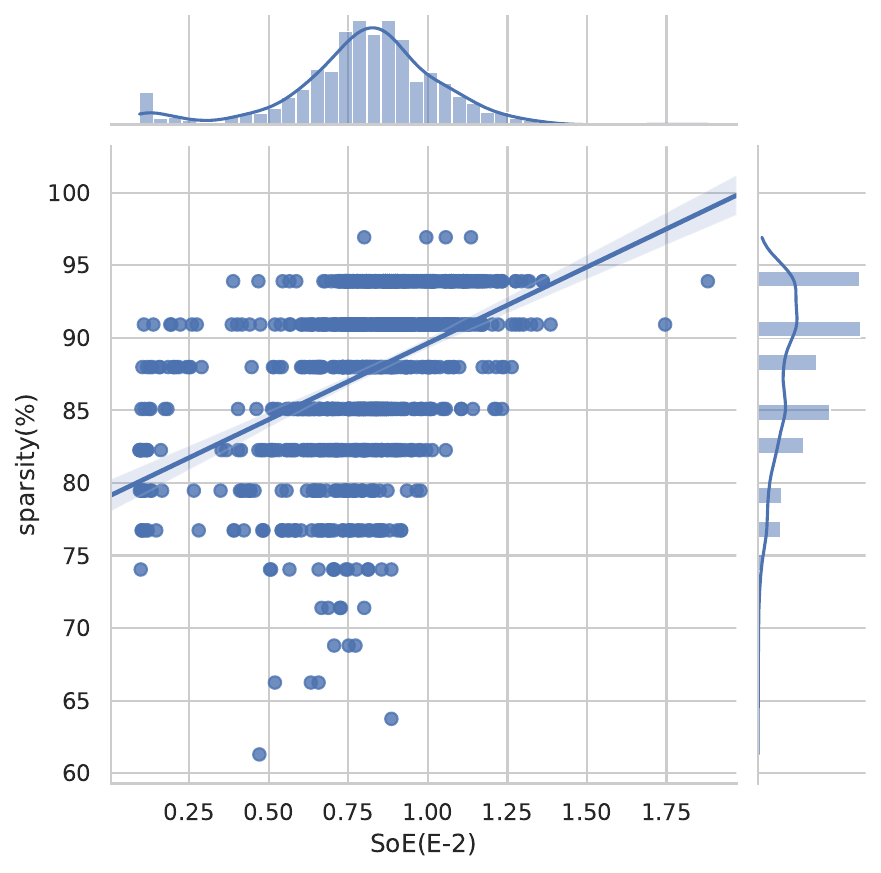}
    \caption{Positive correlation between \name's mask sparsity and head's dependency on position (SoE).}
    \label{fig:sparse_SoE}
\end{figure}

We name $\sigma_p(\mathbb{E}_t[A_{h,i,j}])$ the Standard division of Expectation (SoE) of head $h$.
Note that the expectation is taken over different sentences, while the standard deviation is taken over different attention positions.
Since $\delta_p$ is the same for all heads, we derive that the position impact $\alpha_h$ is proportional to the SoE of different heads.

This conclusion quantifies the observation stated in Section~\ref{sec:mask_vis}. Intuitively, SoE indicates how uneven the mean attention matrix is and thus reflects the influence of position on the attention values. \name's generated mask density shows a positive relationship with SoE, suggesting that \name successfully captures the semantic information of the dense language model, as shown in Figure~\ref{fig:sparse_SoE}.